\documentclass[preprint,journal]{vgtc}            

\onlineid{1761}

\preprinttext{To appear in IEEE Transactions on Visualization and Computer Graphics.}

\vgtccategory{Research}

\title{FADEx: Feature Attribution and Distortion-based Explanation of Dimensionality Reduction}

\author{%
  Lucas Greff Meneses, Evandro S. Ortigossa, Claudio Silva, and Luis Gustavo Nonato
}

\authorfooter{
  \item
  	Lucas Greff Meneses is with University of S\~ao Paulo.
  	E-mail: lucasgreffmeneses@usp.br.
  \item
  	Evandro S. Ortigossa is with Weizmann Institute of Science.
  	E-mail: evandro.scudeleti-ortigossa@weizmann.ac.il.

  \item Claudio T. Silva is with New York University.
  	E-mail: csilva@nyu.edu.

  \item Luis Gustavo Nonato is with University of S\~ao Paulo.
  	E-mail: gnonato@icmc.usp.br.
}

\abstract{%
Dimensionality Reduction (DR) is a fundamental tool for high-dimensional data exploration, reducing the complexity of latent spaces of machine learning models, and assisting in the explanation of complex 
opaque models. However, non-linear DR techniques 
often function as opaque transformations themselves, making it challenging to understand how individual features influence instance positioning in the reduced space. This lack of transparency complicates the analysis and interpretation of structural patterns, hindering the ability to reason about the organization of high-dimensional data based on the projected layout. In order to address this challenge, dimensionality reduction explanation methods have shown promise in improving the understanding of the observed groups and cluster structures. Unfortunately, existing DR explanation approaches tend to suffer from limitations such as multiple attributions per feature and restricted applicability to specific dimensionality reduction methods, which hinder their use. In this work, we propose FADEx, a novel local per-instance feature attribution method that leverages local linear approximation via first-order Taylor expansion and Singular Value Decomposition to provide explanations.
FADEx computes the local linear models via weighted least squares, eliminating the need for out-of-sample data mapping, making it agnostic to the DR method, while simultaneously providing local feature attributions and distortion analysis. Through qualitative and quantitative evaluations, comparisons with existing methods, and case studies, we demonstrate FADEx's effectiveness and versatility in providing explanations and analytical resources for analyzing the behavior of DR methods. The results indicate FADEx yields robust and reliable explanations, outperforming existing approaches in several aspects. 
}


\graphicspath{{figs/}{figures/}{pictures/}{images/}{./}} 

\usepackage{mathptmx}                  
\usepackage{amsmath,amsfonts}
\usepackage{algorithmic}
\usepackage{algorithm}
\usepackage{array}
\usepackage[font=normalsize,labelfont=sf,textfont=sf]{subfig}
\usepackage{multirow,booktabs}
\usepackage{stfloats}
\usepackage{url}
\usepackage{verbatim}
\usepackage{graphicx}
\usepackage{xcolor}
\usepackage{comment}

\usepackage{tikz-cd}
\newcommand{\updated}[1]{#1}
\newcommand{\myhl}[1]{\noindent\textbf{#1}}

\begin{document}

\firstsection{Introduction}

\maketitle

Dimensionality Reduction (DR) has long been a main mechanism for visualizing and exploring high-dimensional data~\cite{nonato2018multidimensional}, supporting the identification of patterns~\cite{espadoto2019toward,ray2021various} and underlying structures within high-dimensional datasets~\cite{guardieiro2025,lee2005nonlinear}. Moreover, DR plays an important role in reducing computational complexity and enhancing the efficiency of machine learning (ML) models~\cite{obaid2019impact,sakr2025espace,thakkar2024fusion}. DR methods also assist in interpreting black-box ML models~\cite{rauber2016visualizing,wang2024visual}, providing insights into learned representations and their impact on model outputs.

Like complex ML models, DR methods, especially non-linear ones, operate as opaque models, making their 
\updated{behavior} difficult to interpret. Moreover, assessing the local influence of individual features on instance positioning in the reduced space is complicated, posing significant challenges in explaining the layouts produced by non-linear DR techniques. Locally evaluating the contribution of each feature to a model's outcome has already proven valuable in the ML domain, where feature attribution techniques have been leveraged for this purpose~\cite{ortigossa2024explainable}. 

Local feature attribution (LFA) refers to a family of methods designed to explain ML models by quantifying how much each input feature contributed to the output for a given instance~\cite{minh2022explainable}. This family of methods produces one of the most granular levels of explainability, which is particularly useful in dimensionality reduction, as one can identify which original features most strongly influence the embedding of each instance. This fine granularity, in turn, clarifies why certain instances are mapped close to together or far apart,  making the embedding space more interpretable. 
Additionally, these detailed explanations can be aggregated at higher levels to explain cluster formation~\cite{yuan2022subplex}, shedding light on why specific groups of points emerge, a challenging problem extensively studied in the visualization field~\cite{cavallo2018clustrophile,jeon2021measuring,xia2022interactive}. 

The potential of local feature attribution in enhancing DR explainability has driven the development of various methods, including surrogate model-based approaches~\cite{marcilio2021explaining,bibal2020explaining},
counterfactual explanations~\cite{artelt2023here}, and gradient-based techniques~\cite{corbugy2024gradient,pmlr-v119-moor20a}. While these methods have proven useful in specific scenarios, they come with limitations that restrict their broader applicability and adoption. For instance, surrogate model-based techniques often produce two or more attributions per feature, complicating interpretability. 
Determining the extent of layout variation caused by feature perturbations is sufficient to qualify a counterfactual is not an easy task. Moreover, surrogate model-based and counterfactuals typically demand that the DR method map out-of-sample (OOS) instances~\cite{sainburg2021parametric,van2009learning}, which is not doable for important techniques such as t-SNE~\cite{van2008visualizing}.
Gradient-based approaches are customized to operate in specific DR methods, demanding model retraining or non-trivial adaptations to operate in different DR techniques.

Beyond identifying which features are important, a comprehensive explanation of DR layouts requires understanding the geometric nature of the transformation, since the feature attribution alone might be insufficient if the underlying projection introduces large distortions. For instance, local distortion accounts for the degree to which the embedding stretches or shrinks neighborhoods relative to the original high-dimensional space. It can lead to misleading interpretations of cluster density and proximity, a problem addressed in several works ~\cite{cashman2025critical,jeon2025unveiling}.

Furthermore, importance scores alone often fail to capture the directional influence of features. Understanding how changes in a specific feature move instances in the visual space provides a more intuitive understanding of the DR behavior. In other words, the directional sensitivity of the DR method allows for a localized understanding of the mapping process~\cite{Ovcharenko_2024}. However, existing explanation techniques often treat attribution and distortion as separate problems, requiring different computational pipelines that may not be easily integrated or applied agnostically across different DR algorithms.

In this work, we introduce \emph{FADEx}, a novel local feature attribution method for explaining dimensionality reduction techniques that unifies explanation, distortion quantification, and directional feature influence. FADEx is designed to address key limitations of existing DR explanation methods while enabling a range of visualization resources. Unlike approaches that rely on neighbor sampling and feature perturbations, which lead to the need to map out-of-sample data, FADEx is based on a local linear approximation derived from a first-order Taylor expansion and Singular Value Decomposition (SVD), eliminating the need for mapping out-of-sample data, ensuring model agnosticism and resulting in a single attribution per feature. Moreover, the proposed methodology naturally incorporates mechanisms to assess distortions and to analyze how the layout would change under directional perturbations. 

We show the effectiveness of FADEx through both qualitative and quantitative comparisons with state-of-the-art explanation methods including LXDR~\cite{bardos2022local}, ClusterShapley~\cite{marcilio2021explaining}, and the gradient-based approach proposed by Corbugy et al.~\cite{corbugy2024gradient}. In addition, we demonstrate its practical utility across a variety of use cases involving different datasets and application scenarios. These results highlight FADEx’s potential to enhance visualization tools that leverage DR techniques for high-dimensional data analysis.

In summary, the main contributions of this work are:

\begin{itemize}
    \item \textbf{FADEx, a novel local feature attribution method for explaining dimensionality reduction techniques.} FADEx leverages local linear approximation via first-order Taylor expansion and SVD to determine feature importance for DR mappings. The use of least squares to approximate the Jacobian matrix eliminates the need for out-of-sample data, making FADEx agnostic to the choice of DR method. Our technique is built on a solid mathematical foundation and addresses many limitations of existing DR explanation techniques.

    \item We evaluate the performance and consistency of FADEx through a series of \textbf{qualitative and quantitative experiments,} comparing its explanatory power with other explanation methods.

    \item We demonstrate the usefulness of FADEx in a set of \textbf{case studies} using diverse datasets. We show the flexibility and usefulness of our approach by using it to explain the behavior of several DR methods, including t-SNE, UMAP, among others.

\end{itemize}

\section{Related Work}

The literature on dimensionality reduction and explainability is extensive. To better contextualize our work and its contribution, we focus the following discussion on techniques specifically designed to interpret/explain DR techniques. For a more comprehensive exploration of both topics, we refer readers to surveys on dimensionality reduction~\cite{espadoto2019toward,jia2022feature,nonato2018multidimensional} and explainability~\cite{burkart2021survey,minh2022explainable,ortigossa2024explainable}.

\smallskip

\myhl{Visualization-Assisted DR Interpretation/Explanation.} A wide range of visualization tools has been developed to help users interpret and explain DR outputs without resorting directly to feature attribution. Instead, they leverage interactive visual interfaces with interconnected views, enabling users to explore DR layouts and locally assess
projection quality~\cite{jeon2021measuring,lespinats2011checkviz,liu2014distortion,martins2015explaining,seifert2010stress}, \updated{cluster formation and the feature contributing to it}~\cite{cavallo2018clustrophile,fujiwara2019supporting,li2022incorporation,thijssen2024interactive}, and feature/axis alignment~\cite{coimbra2016explaining,faust2018dimreader,kwon2017AxiSketcher}. There are also visualization-assisted tools that integrate multiple analytical resources into linked views, enabling the interpretation of \updated{how features influence} the resulting layout~\cite{eckelt2022visual,fujiwara2023feature,montambault2024dimbridge,pagliosa2016understanding,da2015attribute,sohns2021attribute} and the simultaneous analysis of distortions and group formation~\cite{chatzimparmpas2020t,stahnke2015probing}. The visualization-based solutions described above offer visual resources to assist users in interpreting and explaining DR results. However, since they do not rely on feature attribution mechanisms, \updated{which enable quantifying the importance of each feature}, these approaches may \updated{become less intuitive, mainly} for users unfamiliar with traditional ML explanation methods, making it more challenging to understand how specific features influence the mapping process. 

\smallskip

\myhl{Neural Network-Based DR Explanation.} Another approach that has gained prominence in recent years is the development of neural networks (NN) trained to project high-dimensional data into a lower-dimensional space, \updated{which can be organized into two categories. The first category learns a parametric mapping emulating an existing DR technique.} 
Geometry Regularized AutoEncoders~\cite{duque2020extendable}, parametric t-SNE~\cite{van2009learning}, parametric UMAP~\cite{sainburg2021parametric}, DMT-EV~\cite{zang2022dmt}, and Espadoto et al.~\cite{espadoto2020deep} are representatives of this category of methods.
\updated{The second learns an embedding directly via a structure-preserving objective, without referencing an existing DR mapping, such as}
Topological AutoEncoder (TAE)~\cite{pmlr-v119-moor20a} and SSNP~\cite{espadoto2021self}. 
TAE trains an autoencoder directly on the original data to generate a low-dimensional latent representation that serves as the projection. The model optimizes a loss function that minimizes the difference between topological persistence diagrams computed in the original and latent spaces. SSNP also employs an autoencoder but optimizes a loss function that balances reconstruction and classification errors, with the classifier's labels automatically derived through clustering of the original data.

An advantage of NN-based DR methods is that explanations can be derived using gradient-based mechanisms, similar to those employed in traditional ML models~\cite{nielsen2022robust}. However, these methods also have limitations. \updated{Methods that emulate an existing mapping} are constrained to replicating the behavior of the underlying DR method used to generate the training data, thus not being agnostic to the DR technique. \updated{Those that learn an embedding directly} have yet to achieve competitive performance in terms of projection quality.

\smallskip

\myhl{Feature Attribution-based DR Explanation.} Feature attribution methods have traditionally been used to explain machine learning classification and regression models~\cite{ortigossa2024explainable}. The relevance and widespread application of these techniques have motivated their extension to the context of dimensionality reduction. For instance, ClusterShapley~\cite{marcilio2021explaining} builds upon KernelSHAP~\cite{lundberg2017unified} to locally estimate the contribution of each feature to the formation of clusters during the mapping process~\cite{marcilio2021explaining}. The LIME's~\cite{ribeiro2016should} framework of training local linear interpretable surrogate models has also been exploited to explain DR methods~\cite{bardos2022local,bibal2020explaining,mylonas2024exploring}. An issue when adapting traditional ML explanation methods to explain the behavior of DR techniques is that they assign multiple attribution weights per feature (typically one for each projection axis, such as two in a 2D mapping). This complicates interpretability, as different features may have varying levels of influence on each coordinate of an instance in the projection space, making it harder to identify the overall contribution of each feature.
In order to circumvent this issue, Ghosh et al.~\cite{ghosh2020interpretation,ghosh2020visexpres} proposed a methodology that derives feature attributions from Pearson’s correlation between feature distances rather than local linear surrogate models.
Counterfactual explanations~\cite{artelt2023here,bian2021semantic} have also been an alternative to address the multiple attributions per feature issue. However, determining whether feature perturbations lead to meaningful changes in the model’s behavior remains a nontrivial task, making it challenging to determine the extent to which features should be perturbed to generate counterfactuals. Although restricted to Multidimensional Scaling models~\cite{borg2007modern}, Weighted Multidimensional Scaling (WMDS)~\cite{self2018observation} provides an interesting mechanism for feature attribution by assigning a weight to each feature according to semantic interaction with the DR layout. A similar approach is proposed by Cho et al.~\cite{cho2025towards}, which provides two complementary interaction mechanisms for adjusting feature weights: one that allows users to directly explore how weights influence the projections, and another that uses queries to automatically optimize the weights. FeatureClock ~\cite{Ovcharenko_2024} assigns importance to the features fitting local linear models in each data instance. However, it requires multiple linear models per instance, which is computationally demanding.

Combining DR methods and gradient-based explanations has also been a trend. For instance, Han et al.~\cite{han2023explainable} presented a methodology combining WMDS with a visual backpropagation method for generating saliency maps highlighting image features learned during user interaction.
Corbugy et al.~\cite{corbugy2024gradient} proposed a gradient-based feature attribution method designed to explain non-linear DR mappings. This approach closely mirrors techniques used for explaining machine learning models, making it more accessible to users already familiar with feature attribution methods in the ML domain. However, it is highly dependent on the t-SNE loss function, making it challenging to adapt other DR methods.

\smallskip

\updated{
\myhl{Inverse-Projection and Hierarchical Explanation.} Inverse-projection methods aim to learn a mapping from the projection space back to the original high-dimensional space. NNInv~\cite{espadoto2023unprojection}, for instance, reconstructs the inverse of a projection by training a neural network to approximate this mapping, enabling tasks such as gradient-map visualization that reveal regions where the embedding expands or compresses the original space. iHELP~\cite{zeng2022ihelp} hierarchically partitions the layout produced by a non-linear projection and fits a boundary-preserving affine transformation within each partition, supporting a coarse-to-fine interactive analysis of how original features relate to local regions of the embedding. Although both approaches provide valuable resources for interpretability, they do not perform feature attribution and therefore differ from the explanation methods most commonly used in the literature.
}

\smallskip

\myhl{FADEx vs Existing Methods.} The methodology proposed in this work incorporates several traits not simultaneously present in existing approaches. \updated{Built on the spectral structure of a local linear operator derived from the underlying DR method (the estimated Jacobian matrix), FADEx simultaneously delivers per-feature attribution, directional influence vectors, and a distortion measure within a single framework, unifying functionalities that previous methods typically address through separate mechanisms.
} 
FADEx assigns a single importance value to each feature, making it intuitive for users familiar with traditional ML explanation methods and facilitating the interpretation of how individual features influence the DR layout. Moreover, instead of relying on neighbor sampling and feature perturbation, which require out-of-sample mappings, our approach employs a Taylor expansion–based local linear approximation, yielding an optimal local representation from which both attributions and distortion measures are derived, without the need for OOS data, thereby ensuring that FADEx remains model-agnostic.

To highlight how FADEx differs from existing feature attribution approaches, Table \ref{tab:dr_explain_methods} summarizes and compares the main characteristics of major feature attribution methods for DR explanation. Notice that FADEx presents a unique combination of traits that distinguishes it from existing approaches: it assigns a single attribution value per feature, is model agnostic and computationally efficient, supports distortion analysis, and avoids the use of out-of-sample data.

\begin{table*}[ht]
    \centering
    \caption{Characteristics of DR Explanation Methods}
        \begin{tabular}{|l|l|c|c|c|c|}
            \hline
            \textbf{Method} & \textbf{Attribution Type} & \textbf{DR Agnostic} & \textbf{Speed} & \textbf{Distortion} & \textbf{Requires Out-of-Sample}  \\
            \hline

            DimReader~\cite{faust2018dimreader} 
            & Gradient-based 
            & Limited 
            & Slow 
            & No 
            & No \\

            FeatureClock~\cite{Ovcharenko_2024} 
            & Single per feature 
            & Yes 
            & Medium 
            & No 
            & No \\
            
            ClusterShapley~\cite{marcilio2021explaining} 
            & Single per feature
            & Yes 
            & Medium 
            & No 
            & Yes \\
            
            LXDR~\cite{bardos2022local} 
            & Multiple per feature 
            & Limited 
            & Fast 
            & No 
            & Yes \\
            
            Corbugy~\cite{corbugy2024gradient} 
            & Multiple per feature 
            & t-SNE only 
            & Fast 
            & No 
            & No \\
            
            \textbf{FADEx} 
            & \textbf{Single per feature} 
            & \textbf{Yes} 
            & \textbf{Faster} 
            & \textbf{Yes} 
            & \textbf{No} \\
            \hline
        \end{tabular}
    \label{tab:dr_explain_methods}
\end{table*}

\section{The FADEx Method}

In this section, we present the mathematical formulation of our approach. It also describes computational aspects involved in the implementation. 

\subsection{Mathematical Formulation}
\label{sec:math_form}

Let $X$ be a dataset where each $\mathbf{x}=[x_1,\ldots,x_n]\in X$ is an instance in $\mathbb{R}^n$ and $\mathcal{M}:\mathbb{R}^n\rightarrow\mathbb{R}^d,\, d<<n$, be a dimensionality reduction model such that $\mathcal{M}(\mathbf{x})=[m_1(\mathbf{x}),\ldots,m_d(\mathbf{x})]$ ($d=2$ in our context). Assuming that $\mathcal{M}$ is a \updated{local} smooth mapping, that is, the partial derivatives $\frac{\partial^s m_i(x)}{\partial x_j^s}$ exist and are continuous for any $s$, the mapping $\mathcal{M}$ can locally be approximated in a small enough neighborhood of each point $\mathbf{x}\in X$ by the first-order Taylor expansion:
\begin{equation}
    \mathcal{M}(\mathbf{x+h})\approx \mathcal{M}(\mathbf{x})+J_{\mathcal{M}(\mathbf{x})}\cdot \mathbf{h}
    \label{eq:taylor}
\end{equation}
where $J_{\mathcal{M}(\mathbf{x})}$ is the $d\times n$ \textit{Jacobian matrix} (linear transformation) of $\mathcal{M}$ in $\mathbf{x}$, given by:
\begin{equation}
J_{\mathcal{M}(\mathbf{x})}=\begin{bmatrix}
\frac{\partial m_1(\mathbf{x})}{\partial x_1} & \cdots & \frac{\partial m_1(\mathbf{x})}{\partial x_n}\\
 & \vdots & \\
 \frac{\partial m_d(\mathbf{x})}{\partial x_1} & \cdots & \frac{\partial m_d(\mathbf{x})}{\partial x_n}
\end{bmatrix}
\label{eq:jacobian}
\end{equation}

\updated{

Since explicit expressions for these partial derivatives are generally unavailable for methods such as t-SNE and UMAP, FADEx does not compute $J_{\mathcal{M}(\mathbf{x})}$ analytically. Instead, it estimates an approximated Jacobian in the local neighborhood of each instance (see \cref{sec:computational_aspects}). Therefore, it is required that the underlying DR method is locally smooth to enable a good approximation of  
$\mathcal{M}$ by a linear operator within a small neighborhood of each instance. 

}

The linear transformation $J_{\mathcal{M}(\mathbf{x})}$ provides a local approximation of $\mathcal{M}$ within a sufficiently small neighborhood around $\mathbf{x}$, governed by $\mathbf{h}$. As a linear transformation, $J_{\mathcal{M}(\mathbf{x})}$ can be expressed using Singular Value Decomposition (SVD) in the form: 
\begin{equation}
    J_{\mathcal{M}(\mathbf{x})}=U\Sigma V^\top
    \label{eq:svd}
\end{equation}
where $U$ and $V$ are $d\times d$ and $n\times d$ matrices, respectively, with orthonormal column vectors  (the symbol $\top$ means transpose). $\Sigma$ is a $d\times d$ diagonal matrix whose diagonal entries are non-negative values called \emph{singular values}~\cite{kalman1996singularly}.

The SVD decomposition describes how the linear approximation $J_{\mathcal{M}(\mathbf{x})}$ locally maps instances from $\mathbb{R}^n$ to $\mathbb{R}^d$. The annotated equation~\cref{eq:svd_a} shows how $J_{\mathcal{M}(\mathbf{x})}$ operates in a neighborhood of $\mathbf{x}$:
\begin{equation}
J_{\mathcal{M}(\mathbf{x})}\cdot \mathbf{h}=\Bigl(\underbrace{U\bigl(\overbrace{\Sigma(\underbrace{V^\top \mathbf{h}}_{\text{projection}}}^{\text{scale}})\bigl)}_{\text{point in}\,\mathbb{R}^d}\Bigl)
\label{eq:svd_a}
\end{equation}

Denoting the columns of $U$ and $V$ as $\{\mathbf{u}_1,\dots,\mathbf{u}_d\}$ and $\{\mathbf{v}_1,\dots,\mathbf{v}_d\}$ respectively,
and the ordered singular values in $\Sigma$ as $\lambda_1\geq\lambda_2\geq\ldots\geq\lambda_k$, \cref{eq:svd_a} becomes:
\begin{equation}
J_{\mathcal{M}(\mathbf{x})}\cdot \mathbf{h}=(\lambda_1 \mathbf{v}_1^\top \mathbf{h})
\begin{bmatrix}
| \\ \mathbf{u}_1 \\ |
\end{bmatrix}+\cdots+
(\lambda_d \mathbf{v}_d^\top \mathbf{h})
\begin{bmatrix}
| \\ \mathbf{u}_d \\ |
\end{bmatrix}
\label{eq:svd_expansion1}
\end{equation}
where $\mathbf{u}_i$ and $\mathbf{v}_i$, $i=1,\ldots,d$, are $d$ and $n$-dimensional vectors, respectively, and $\mathbf{v}_i^\top \mathbf{h}$ corresponds to the dot product between $\mathbf{v}_i$ and $\mathbf{h}$.

From \cref{eq:svd_a} and \cref{eq:svd_expansion1}, it follows that the local linear mapping $J_{\mathcal{M}(\mathbf{x})}$ first projects $\mathbf{h}$ onto a $d$-dimensional subspace $S\subset\mathbb{R}^n$, spanned by the orthonormal basis $\{\mathbf{v}_1,\dots,\mathbf{v}_d\}$. The resulting projections, given by $\mathbf{v}_j^\top \mathbf{h}$, are then scaled by the corresponding singular values, yielding the terms $(\lambda_i \mathbf{v}_i^\top \mathbf{h})$. These values are used as coefficients in the linear combination of the vectors $\mathbf{u}_i$, ultimately generating a point in $\mathbb{R}^d$.

If $\mathbf{h}$ is aligned with the canonical coordinate axis \( \mathbf{e}_j \), meaning \( \mathbf{h} = \mathbf{h}_j = [0, \ldots, 0, h_j, 0, \ldots, 0] \), then the coefficients \( (\lambda_i \mathbf{v}_i^\top \mathbf{h}) \) in the linear combination given in \cref{eq:svd_expansion1} simplify to \( (\lambda_i v_{ij} h_j) \), where \( v_{ij} \) represents the \( j \)-th coordinate of \( \mathbf{v}_i \). 
Consequently, \emph{larger} values of \( (\lambda_i v_{ij} h_j) \) indicate a \emph{greater} contribution of the \( j \)-th feature to the resulting point in \( \mathbb{R}^d \).  
Building on this observation, given that the singular values are ordered in descending order, with \( \lambda_1 \geq \lambda_2 \geq \dots \geq \lambda_d \), we set \( h_j = x_j \), where $x_j$ is the \( j \)-th feature of \( \mathbf{x} \). This assignment provides a natural ordering on the features according to their importance.
From this scenario, we define the importance/attribution of each feature $j$ for the mapping of instances from a neighborhood of $\mathbf{x}$ to $\mathbb{R}^d$ as follows:
\begin{equation}
    \phi_j = \sum_{i=1}^d  \frac{\lambda_i}{\lambda_1} \left|v_{ij} x_j\right|,\,\, j = 1, \dots, n 
    \label{eq:feat_importance}
\end{equation}
where $|\cdot|$ is the absolute value and the ratio $\frac{\lambda_i}{\lambda_1}$ is just a scaling factor. When $d=2$, \cref{eq:feat_importance} becomes simply $\phi_j = \left|v_{1j} x_j\right|+\frac{\lambda_2}{\lambda_1} \left|v_{2j} x_j\right|$. 

\smallskip

\myhl{Local Attribution Computation.} The mathematical formulation above results in attributions computed locally for each instance $\mathbf{x}$, that is, the 
\updated{Jacobian ${J}_{\mathcal{M}(\mathbf{x})}$ (in fact, an approximation version of it, as described in \cref{sec:computational_aspects})}
and the SVD decomposition are computed in the local neighborhood of each point. Therefore, different regions of the data space can exhibit different feature importance patterns.

Moreover, an attribution value is assigned to each feature of each instance \( \mathbf{x} \in X \). Therefore, the feature attribution process can be interpreted as a mapping 
\begin{equation}
\begin{split} 
    \Phi_{\mathcal{M}}: X \subset \mathbb{R}^n \to E\subset\mathbb{R}^n, \\
    \Phi_{\mathcal{M}}(\mathbf{x}) = [\phi_1(\mathbf{x}), \ldots, \phi_n(\mathbf{x})]       
\end{split}
    \label{eq:expl_space}
\end{equation}

\updated{where $E$ denotes the \emph{explanation space}, in which each instance $\mathbf{x}$ is represented by its attribution vector $\Phi_{\mathcal{M}}(\mathbf{x})$. The $j$-th coordinate of this vector quantifies the importance of feature $j$ for the DR mapping of $\mathbf{x}$.}

\begin{figure}[t]
    \centering
    \includegraphics[width=0.7\linewidth]{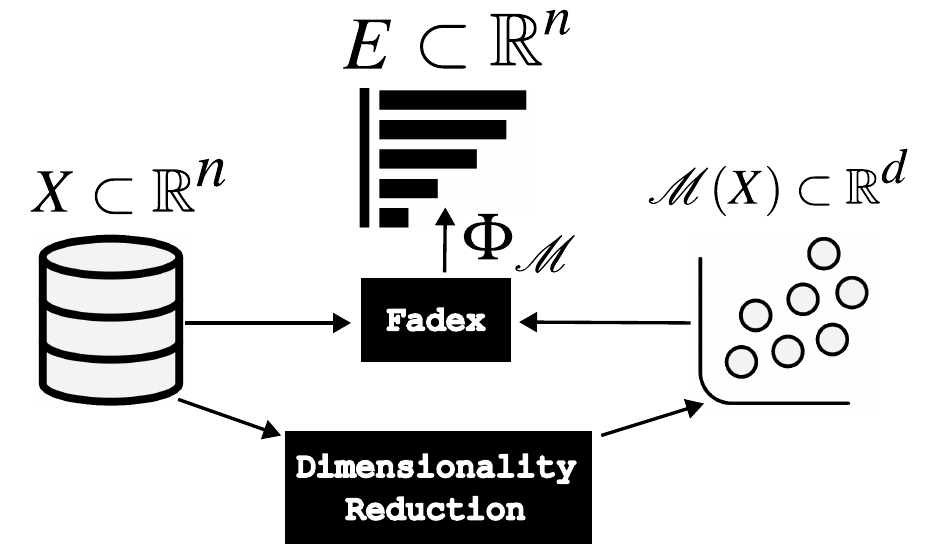}
    \caption{The dimensionality reduction method transforms instances from the data space into the projection space. FADEx then takes both the original and projected instances as input, producing feature attributions and distortion measures that can be leveraged for visualization and interpretation.}
    \label{fig:explanation}
\end{figure}

\myhl{Intuitive Explanation.} In less mathematical terms, each instance $\mathbf{x}$ is associated with an \emph{attribution vector} in the explanation space $E$. The quantity 
$\phi_i(\mathbf{x})$  reflects the importance of the $i$-th feature for the dimensionality reduction process within the local neighborhood of \( \mathbf{x} \).  
The attribution vectors can be explored in different ways (see \cref{sec:cs}) to generate meaningful information visualized in the projection space. \cref{fig:explanation} illustrates the whole explanation process.

\smallskip

\myhl{Feature Influence Vectors.} In addition to enabling the computation of feature attributions $\phi_i$, the 
\updated{(estimated) Jacobian $\widehat{J}_{\mathcal{M}(\mathbf{x})}$}
also provides directional information. For each feature $x_i$, we define the \textit{Feature Influence Vector} $g_i(\mathbf{x}) \in \mathbb{R}^d$ as the $i$-th column of ${J}_{\mathcal{M}(\mathbf{x})}$. When $d=2$, these vectors can be plotted in the projected space, indicating the direction in which the output moves when
$\mathbf{x}$ is slightly perturbed along the $x_i$ direction.

\smallskip

\myhl{Mapping Distortions.} The spectrum of the \updated{estimated} Jacobian matrix provides a valuable tool for assessing how $\mathcal{M}$ distorts the neighborhood around each instance $\mathbf{x}$. Specifically, the largest singular value, $\lambda_1$, corresponds to the spectral norm of 
\updated{$\widehat{J}_{\mathcal{M}(\mathbf{x})}$}
and indicates the extent of local stretching ($\lambda_1>1$) or shrinking ($\lambda_1<1$) along the direction of the corresponding singular vector $\mathbf{u}_1$. Thus, color-coding points in the DR layout according to \( \lambda_1 \) enables visual identification of regions where the mapping induces expansion or compression. We denote the Jacobian Spectral Norm distortion measure as \emph{SND}.

\subsection{Approximating the Jacobian Matrix}
\label{sec:computational_aspects}

An important component of the FADEx formulation is the estimation of the Jacobian matrix, as explicit mathematical expressions for the derivative of the coordinate functions $m_i(\mathbf{x}),\, i=1,\ldots,d$ (see \cref{eq:jacobian}) are typically difficult to obtain.

To avoid the use of out-of-sample data, which several DR methods cannot handle, we estimate $J_{\mathcal{M}(\mathbf{x})}$ from the local geometry of the original and mapped instances, fitting a local linear model via weighted least squares. Given an instance $\mathbf{x} \in X$ and its embedding $\mathbf{y} = \mathcal{M}(\mathbf{x})$, let $\mathcal{N}_{\mathbf{x}} = \{ (\mathbf{x}_\ell, \mathbf{y}_\ell)\}_{\ell=1}^{k}$ be the set of $k$-nearest neighbors $\{\mathbf{x}_\ell\}$ of $\mathbf{x}$ in the high-dimensional space, together with their corresponding low-dimensional mappings $\{\mathbf{y}_\ell\}$. Assuming $\mathcal{M}$ is locally smooth, we approximate the first-order Taylor relation (\cref{eq:taylor}) by:

\begin{equation}
    \mathbf{y}_\ell - \mathbf{y} \approx J_{\mathcal{M}(\mathbf{x})} (\mathbf{x}_\ell - \mathbf{x}),\quad \ell=1, \dots, k
    \label{eq:taylor2}
\end{equation}

Stacking the centered displacements into matrices

\begin{equation}
\begin{aligned}
\Delta X &\in \mathbb{R}^{k\times n},
&\qquad \Delta X_{\ell,:} &= (\mathbf{x}_\ell-\mathbf{x})^\top, \\
\Delta Y &\in \mathbb{R}^{k\times d},
&\qquad \Delta Y_{\ell,:} &= (\mathbf{y}_\ell-\mathbf{y})^\top ,
\end{aligned}
\end{equation}
from \cref{eq:taylor2} we obtain $\Delta Y \approx \Delta X B$, where 
$J_{\mathcal{M}(\mathbf{x})} = B^\top$.

To emphasize proximity, we weight each neighbor by its high-dimensional distance $r_\ell = \| \mathbf{x}_\ell - \mathbf{x} \|$ using a Gaussian kernel,

\begin{equation}
    w_\ell = \exp{\frac{-r_\ell^2}{2\sigma^2}}
    \label{eq:gaussian_weights}
\end{equation}
where $\sigma$ is a local scale parameter (we set $\sigma$ to the median of $\{r_\ell\}_{\ell = 1}^k$). 
\updated{This Gaussian weighting enforces the locality of the Taylor approximation in ~\cref{eq:taylor2}, which is valid only near $\mathbf{x}$. Since distant neighbors may introduce larger approximation errors, reducing their influence ensures that the fit is primarily determined by neighbors for which the approximation remains reliable.} 
The matrix $B$ (and thus the Jacobian matrix) is computed by solving the weighted ridge least-squares problem:

\begin{equation}
\widehat{B}
\;=\;
\arg\min_{B\in\mathbb{R}^{n\times d}}
\bigl\|W^{1/2}(\Delta X B-\Delta Y)\bigr\|_F^2
\;+\;
\alpha \|B\|_F^2,
\label{eq:ridge_wls}
\end{equation}
where $\| \cdot \|_F$ denotes the Frobenius norm, $W$ is the diagonal weight matrix with elements given by \cref{eq:gaussian_weights}, and $\alpha > 0$ is the regularization parameter. The solution of~\cref{eq:ridge_wls} is given by solving the linear system:

\begin{equation}
(\Delta X^\top W \Delta X + \alpha I)\,\widehat{B}
\;=\;
\Delta X^\top W \Delta Y,
\label{eq:normal_eq_ls}
\end{equation}
where $I$ is the identity matrix. Therefore, the estimated Jacobian used by FADEx is:

\begin{equation}
    \widehat{J}_{\mathcal{M}(\mathbf{x})} \approx \widehat{B}^\top
\end{equation}

\smallskip

The ridge regularization term in \cref{eq:ridge_wls} ($\|B\|_F^2$) is used to guarantee numerical stability, as the neighborhood of each instance might be unevenly sampled, making the linear system ill-conditioned. Moreover, the regularization can be interpreted as imposing a prior that large Jacobian entries are unlikely, i.e., local derivatives are expected to be moderate, which is often consistent with the assumption of locally smooth mappings. In our implementation, $\alpha=0.1$ by default, although it is a parameter that can be adjusted in the FADEx framework.

\smallskip

\myhl{The Curse of Dimensionality.} In high-dimensional spaces, data points tend to be widely dispersed, weakening notions of proximity. Consequently, neighborhoods built from nearest neighbors may not be truly local, which can introduce numerical instability not fully mitigated by regularization. To address this issue, we apply two preprocessing steps. First, we discard features with near-zero local variance \updated{(below a threshold of $10^{-6}$)}, as they contribute little to the neighborhood geometry. Second, we perform a local PCA to reduce dimensionality while being conservative to preserve $95\%$ of the explained variance. The Jacobian is estimated in the reduced PCA space and then ``mapped'' to the original feature space using the chain rule for derivatives. Details on how to compute the Jacobian in the original high-dimensional space from its counterpart in the PCA space are provided in the supplementary material. In practice, PCA is applied only when the data dimensionality exceeds $50$, as our experiments indicate that the numerical approximation remains robust for lower-dimensional data.

\smallskip

\myhl{Implementation Details.} The pseudocode detailing all steps of the FADEx feature attribution computation is provided in the supplementary material. FADEx is implemented as a Python package, which is available as open source at \url{https://github.com/greffao/fadex}. 

\subsection{Approximation Quality}

In the following, we assess how accurately the locally approximated Jacobian captures the DR mapping. We conducted an empirical study across several datasets and DR methods (t-SNE~\cite{van2008visualizing}, UMAP~\cite{mcinnes2018umap}, Isomap~\cite{tenenbaum2000global}, and LLE~\cite{roweis2000nonlinear}). Because the embedding space has no intrinsic physical scale, absolute error measures such as the Mean Squared Error between the DR and Jacobian-projected points are not well suited for evaluation, as the magnitude of the error lacks a clear interpretation. Instead, we assess the quality of the approximation by measuring how well it preserves the relative positions of neighboring points.

\smallskip

\myhl{Evaluation Protocol.} For each instance $\mathbf{x}$, we identify its $k$-nearest neighbors in the high-dimensional space and perform a local hold-out validation to test the Jacobian's predictive power. Specifically, we use $70\%$ of the neighbors to estimate the Jacobian $\widehat{J}_{\mathcal{M}(\mathbf{x})}$, reserving the remaining $30\%$ as a test set. For each test neighbor $\mathbf{x}_\ell$, we compute the actual displacement $\Delta \mathbf{y}_\ell = \mathcal{M}(\mathbf{x}_\ell) - \mathcal{M}(\mathbf{x})$ and compare it with the linear approximation $\widehat{\Delta \mathbf{y}}_\ell = \widehat{J}_{\mathcal{M}(\mathbf{x})} (\mathbf{x}_\ell - \mathbf{x})$ by calculating their \textit{Mean Cosine Similarity}. This scale-invariant metric assesses the angular alignment between the predicted and actual displacement vectors. A value near $1.0$ confirms that the approximated Jacobian correctly identifies the direction in which the data should be placed in the 2D layout. We performed this protocol across several datasets, DR methods, DR method parameters, and number of neighbors used to compute $\widehat{J}_{\mathcal{M}(\mathbf{x})}$.

\smallskip

\begin{figure}[tb]
    \centering
    \includegraphics[width=\columnwidth]{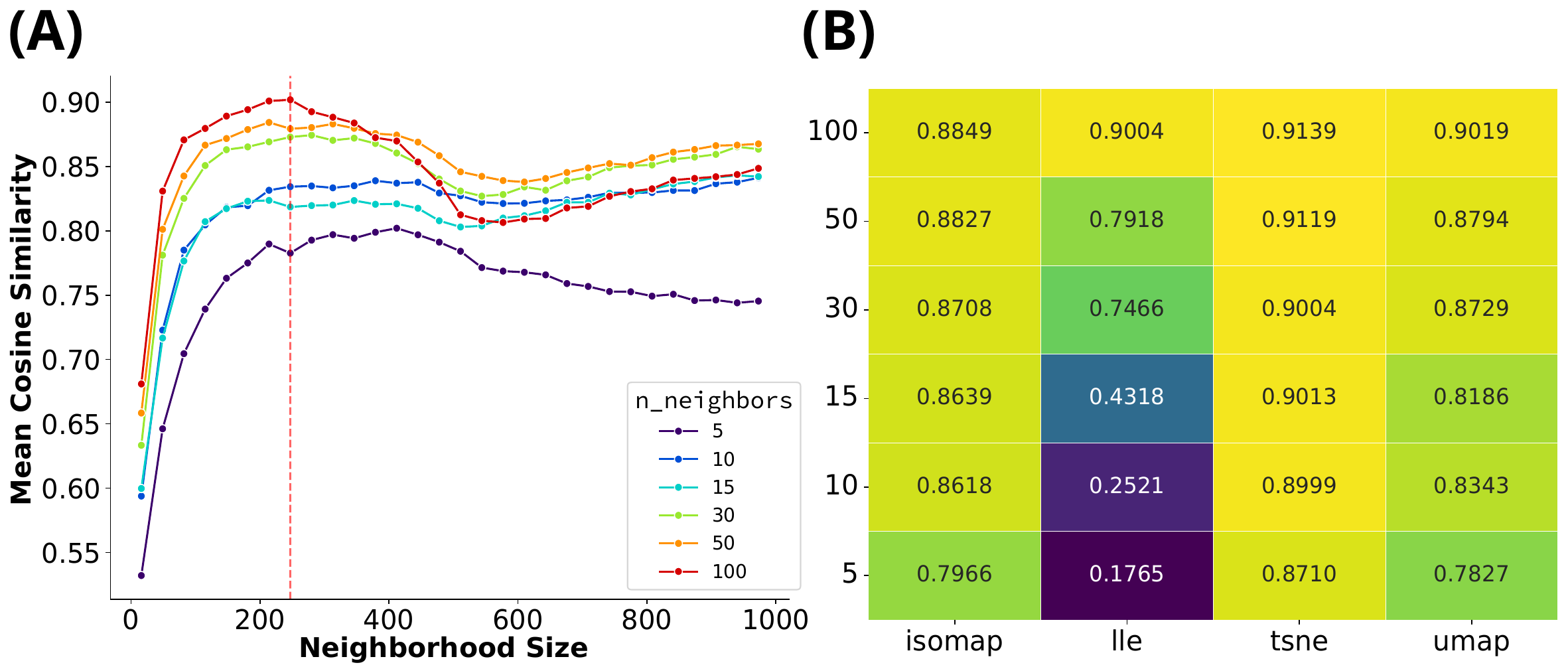}
    \caption{Approximation quality of the local Jacobian for the \textit{Musk Version 2} dataset. (A) shows the variation of the mean cosine similarity across different neighborhood sizes and \texttt{n\_neighbors} parameters for the UMAP method. (B) presents the mean cosine similarity for different DR methods and parameters, using an optimum number of neighbors (vertical red line in (A)).}
    \label{fig:approx_quality}
\end{figure}

\myhl{Results.} \cref{fig:approx_quality} shows the results for a $2000$-sample version of the \textit{Musk Version 2} dataset ($166$-dimensional data), available in \href{https://archive.ics.uci.edu/dataset/75/musk+version+2}{UCI ML Repository}. Figure~\ref{fig:approx_quality}A) shows that, with an appropriate number of neighbors, the approximation achieves a high cosine similarity. Our experiments on several datasets indicate that setting the number of neighbors to roughly $10\%$ of the dataset size ($247$ in the \textit{Musk Version 2} case) generally yields good results.

The heatmap in \cref{fig:approx_quality}B) demonstrates that high approximation fidelity is achievable across various DR methods with different parameters (the numbers on the vertical axis correspond to the number of neighbors parameter for Isomap, LLE, and UMAP, referring to perplexity for t-SNE). As expected, the quality of the Jacobian estimation is directly influenced by the hyperparameters of the DR algorithm itself. Higher values for these parameters generally lead to better approximation results, as they force the DR method to produce a smoother embeddings, which makes the first-order Taylor approximation more consistent. \updated{A more comprehensive assessment of the approximation quality is provided in the supplementary material.}

\section{Validation and Comparisons}
\label{sec:validation}

In this section, we present a series of experiments to validate the correctness of FADEx and build intuition about its analytical capabilities. We also compare FADEx with other state-of-the-art feature attribution methods.

\subsection{Local Explanation Validation}

As a local explanation method, FADEx assigns high relevance to features that are responsible for driving the positioning of points in the low-dimensional embedding. To validate this capability in a controlled setting, we designed a synthetic dataset with known ground-truth feature relevance patterns and evaluated whether FADEx is able to recover these patterns.

We generated a dataset in $12$ dimensions containing three imbalanced clusters with $250$, $500$ and $1250$ samples each. The data was initialized from a uniform distribution in the interval $[0,1]$, defining disjoint clusters by modifying four features per cluster through \updated{a mean offset} ($\mu = 2$) with standard Gaussian noise \updated{($\sigma = 1$)}.
Concretely, cluster $C_0$ was shifted in features $\{0, 1, 2, 3\}$, cluster $C_1$ in features $\{4, 5, 6, 7\}$, \updated{and cluster $C_2$ in features $\{8, 9, 10, 11\}$.
The offset is positive for $C_0$ and $C_1$ and negative for $C_2$, that is, the corresponding features of $C_2$ are shifted towards $-\mu$.

}

\begin{figure}[t]
  \centering
  \includegraphics[width=\columnwidth]{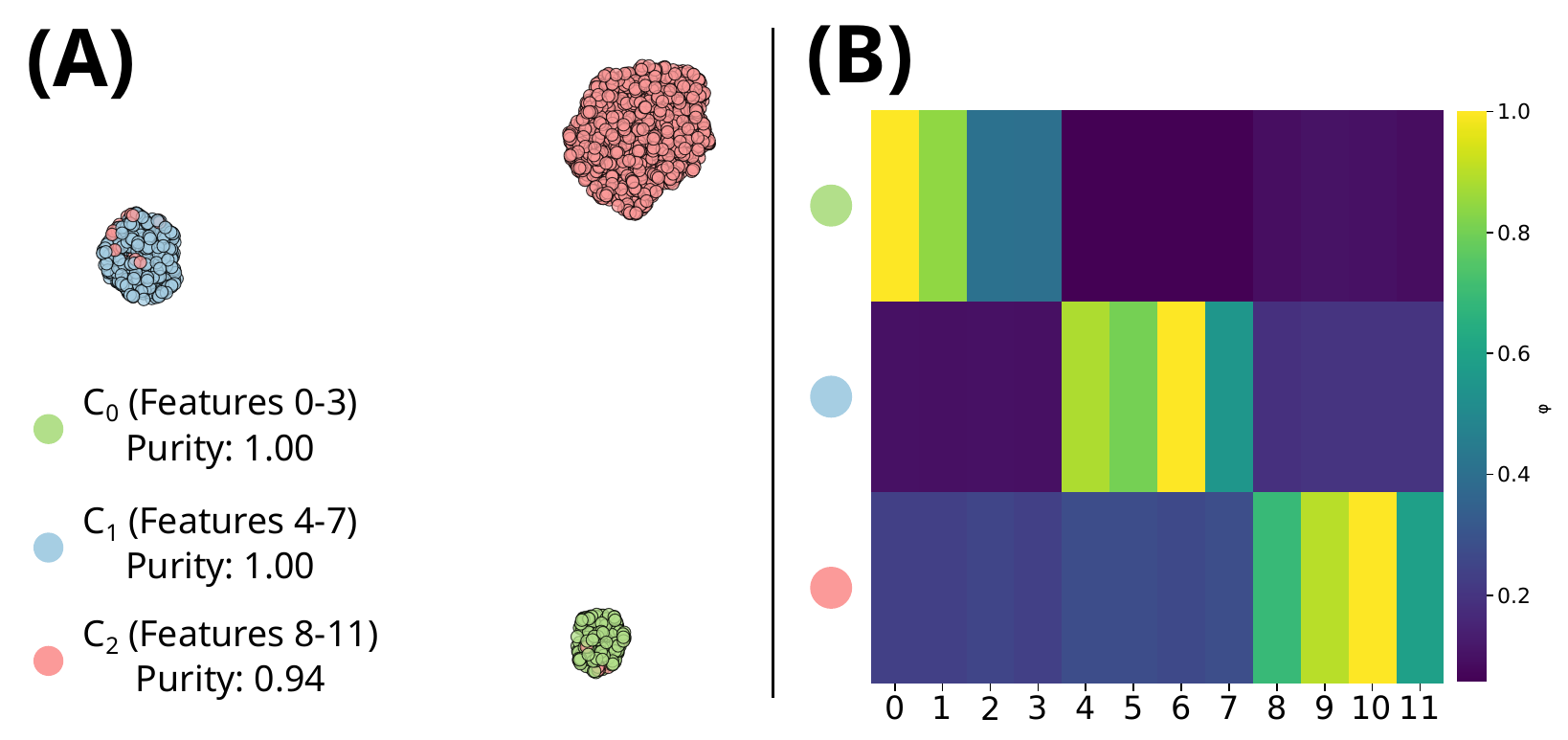}
  
  \caption{FADEx explanations over a synthetic dataset embedded by UMAP. (A) UMAP correctly separated the data into three regions. The point colors correspond to cluster IDs computed in the explanation space $E$ ($k$-means, $k=3$). (B) FADEx individual attribution vectors are aggregated in a heatmap, showing the method's capability to identify the important features in each cluster.}
  \label{fig:local_explanation_validation}
\end{figure}

The data was embedded into two-dimensional space using UMAP, with $15$ neighbors and all other parameters set to default. ~\cref{fig:local_explanation_validation}A) shows the embedding space produced. We applied FADEx to every point in the dataset using $20$ neighbors ($10\%$ of the data). We cluster the attribution vectors of each instance in the explanation space $E$ (\cref{eq:expl_space}) using $k$-means with $k=3$ (chosen after a silhouette score analysis) to gain insights into how explanations differ across the dataset. The cluster colors in~\cref{fig:local_explanation_validation}A) correspond to the cluster IDs obtained in the explanation space, clearly showing three distinct clusters defined by the similarity of the feature attribution values. The \emph{Purity} values in the legend reveal a clear cluster formation, with a small deviation in the larger cluster. 
\cref{fig:local_explanation_validation}B) presents the average feature importance heatmap in each cluster. We computed the mean explanation vector for all points belonging to each identified cluster in the explanation space. 
The resulting heatmap reveals a clear diagonal block structure, aligning perfectly with the ground-truth shifted features. \updated{Note that, since the per-instance noise has unit variance ($\sigma = 1$) relative to a feature offset of magnitude $2$, the clusters generated by shifting the subset of features might still partially overlap with the other clusters.
This leads to the observed unequal feature importance values among the shifted features that characterize the clusters, but the diagonal block structure remains clearly identifiable.} 

\subsection{Validating the SND Metric}
 
Dimensionality reduction methods inherently introduce geometric distortions when embedding data in a low-dimensional space. Identifying where these distortions are most pronounced is crucial for assessing where the mapping is more (or less) reliable.~\cite{lespinats2011checkviz}. 

\smallskip

\myhl{SND vs d/D.} As discussed in the last paragraph in~\cref{sec:math_form}, FADEx can assess mapping distortions using the SND metric. To evaluate SND effectiveness in capturing distortions, we applied t-SNE (\texttt{perplexity=50}) and UMAP (\texttt{n\_neighbors=50}) on different datasets (see~\cref{tab:snd_tnc_corr}), computing the SND metric in a random sample of $500$ instances from each dataset. Using the selected instances, we compute the well-known \emph{Distance Preservation Ratio} $\delta = d/D$ distortion metric, where $D$ and $d$ are the local average distances in the high and low-dimensional spaces, respectively. This metric has been chosen because, like SND, it also captures "compression" or "expansion" of distances during the dimensionality reduction process, different from other metrics such as Trustworthiness and Continuity that leverages neighbors' ranking, thus not being appropriate for a fair comparison with SND. To address the different scales in high- and low-dimensional spaces, each \updated{feature} is normalized to the interval $[0, 1]$ in the original and reduced spaces to share a common range.

We quantitatively compare SND and $\delta$ by examining the correlation between the two metrics. \cref{tab:snd_tnc_corr} reports these correlations across different datasets embedded in 2D using t-SNE and UMAP. 

Overall, the results show a good agreement between the two metrics. For t-SNE, correlations are moderate ($0.3$ – $0.6$) in three out of seven datasets and high in the remaining four. For UMAP, correlations are high in five datasets ($0.7$-$1.0)$ and moderate in the other two. Despite their distinct mathematical and computational foundations, SND and $\delta$ exhibit consistent behavior, indicating that SND is capable of properly capturing distortion patterns. 

\begin{table}[t]
    \centering
    \caption{Correlation between SND and the local distance ratio $\delta=d/D$ for t-SNE and UMAP across different datasets. $\delta$ measures the ratio between mean embedding distance $d$ to the mean original distance $D$ over each point's $k$-nearest neighborhood. Correlation confirms a good agreement between SND and $\delta$.}
    \label{tab:snd_tnc_corr}
    \small
    \setlength{\tabcolsep}{10pt} 
    \renewcommand{\arraystretch}{1.3}
    
    \begin{tabular}{|l|c|c|c|}
        \hline
        \textbf{Dataset} & \textbf{Dimensions ($N \times D$)} & \textbf{t-SNE} & \textbf{UMAP} \\
        \hline
        \href{https://scikit-learn.org/stable/modules/generated/sklearn.datasets.load_wine.html}{Wine}             & $178 \times 13$ & $0.86$ & $0.70$\\
        
        \href{https://archive.ics.uci.edu/dataset/17/breast+cancer+wisconsin+diagnostic}{Breast Cancer}   & $569 \times 30$ & $0.92$ & $0.89$ \\
        
        \href{https://archive.ics.uci.edu/dataset/33/dermatology}{Dermatology}       & $366 \times 34$ & $0.86$ & $0.81$\\
        
        \href{https://archive.ics.uci.edu/dataset/602/dry+bean+dataset}{Dry Bean}         & $13611 \times 16$ & $0.90$ & $0.92$\\
        
        \href{https://archive.ics.uci.edu/dataset/75/musk+version+2}{Musk Version 2}    & $6598 \times 166$ & $0.37$ & $0.78$\\
        
        \href{https://archive.ics.uci.edu/dataset/54/isolet}{Isolet}             & $7797 \times 617$ & $0.63$ & $0.56$\\
        
        \href{https://scikit-learn.org/stable/modules/generated/sklearn.datasets.fetch_openml.html}{MNIST (sampled)}                & $5000 \times 784$ & $0.60$ & $0.69$\\
        
        \hline
    \end{tabular}
\end{table}

\smallskip

\myhl{SND and Nearest Neighbors.} As previously discussed, the SND metric captures distortions in two complementary ways: values below one indicate neighborhood shrinking, while values above one indicate expansion, and closer to $1$ correspond to lower distortion.

\begin{figure}[t]
  \centering
  \includegraphics[width=\columnwidth]{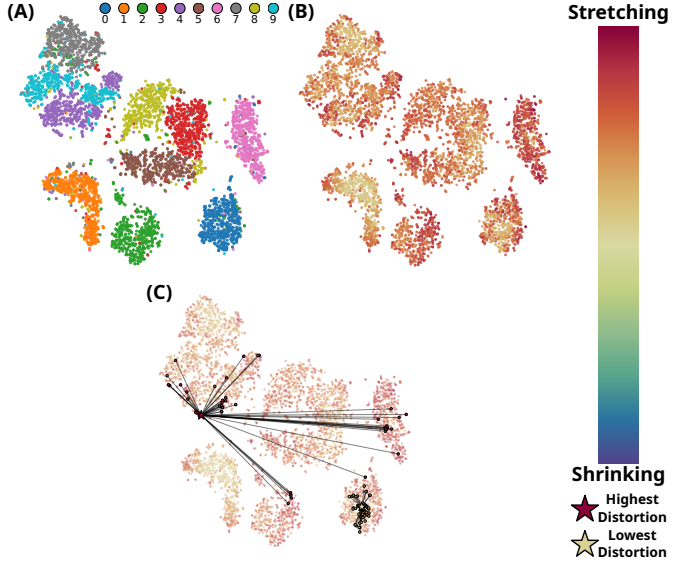}
  
  \caption{FADEx's SND distortion visualization and its relation with neighborhood structures. (A) MNIST low-dimensional embedding created by t-SNE. (B) SND distortion map indicating that distortions arise from neighborhood stretching. (C) For a high- and a low-distortion point, their $k$-nearest neighbors in the original high-dimensional space, plotted in the 2D embedding.}
  \label{fig:snd_hypertrix}
\end{figure}

To validate whether SND effectively reflects this behavior, we conducted the qualitative analysis shown in~\cref{fig:snd_hypertrix}. We applied t-SNE (\texttt{perplexity}=30, \texttt{early\_exaggeration}=24, \texttt{learning\_rate}=600, \texttt{init}=’pca’) to $5000$ randomly sampled instances from the MNIST dataset (\cref{fig:snd_hypertrix}A) and computed the corresponding SND values. The resulting distortion map (\cref{fig:snd_hypertrix}B), visualized with a diverging colormap (dark blue and dark red associated with neighborhood shrinking and stretching, respectively), shows that most distortions arise from neighborhood expansion, indicating that t-SNE tends to stretch local structures.

To further assess the validity of SND, \updated{\cref{fig:snd_hypertrix}C shows the $k$-nearest neighbors in the original high-dimensional space, plotted in the 2D embedding, for two points: one with high and another with low distortion. The neighbors of the high-distortion point are widely dispersed across the 2D layout, whereas those of the low-distortion point remain locally arranged.} This comparison confirms that SND effectively captures the distortion introduced during the mapping process.

\subsection{Comparing FADEx with other Explanation Methods}

In the following, we compare FADEx with three DR explanation methods:  Corbugy~\cite{corbugy2024gradient}, LXDR~\cite{bardos2022local}, and ClusterShapley~\cite{marcilio2021explaining}. These approaches have been chosen because, like FADEx, they are feature-attribution-based methods grounded in distinct mathematical and computational frameworks, leading to different trade-offs in terms of computational cost, robustness, and the semantic interpretation of their explanations.

Corbugy’s explanation method~\cite{corbugy2024gradient} builds on the analytical derivatives of the Kullback–Leibler divergence loss driving the t-SNE embeddings. These derivatives yield a matrix in which the magnitude of the $i$-th column reflects the contribution of the $i$-th feature to the embedding. The analytical derivatives ensure numerical stability; however, extending the approach to DR techniques beyond t-SNE is not straightforward. 
LXDR~\cite{bardos2022local} fits a local linear surrogate model per embedding dimension, thus assigning an importance value per embedded dimension (two per feature). It relies on sampling and projecting a neighborhood, which requires out-of-sample mapping support. While model-agnostic, its attributions reflect the surrogate model rather than the original DR mapping.
ClusterShapley explains cluster formation rather than individual embeddings, using predefined clusters to estimate membership probabilities and compute Shapley values that quantify each feature’s contribution to each cluster assignment, therefore making multiple attributions per feature (one for each cluster).

FADEx exhibits key structural differences relative to the methods above. Unlike LXDR, it does not require out-of-sample data and assigns a single attribution per feature. In contrast to Corbugy’s method, FADEx is model-agnostic. While ClusterShapley focuses on explaining cluster formation, FADEx targets the local behavior of the DR mapping, identifying which features drive it at the instance level. 
\updated{Moreover, as depicted in Table 1 of the supplementary material, FADEx is nearly one order of magnitude faster than the other three methods, while presenting lower memory consumption, mainly in larger datasets. Sec. 5 of the supplementary material also includes a discussion of FADEx's visualization-encoding choices and how they relate to the visual encodings of the other explanation methods.}

\subsubsection{Ablation Test}

Ablation is a common approach to evaluating whether top-ranked features are genuinely the most important to a ML model's decision~\cite{hameed2022based}. This process involves identifying the top-$k$ most important features, removing them, and then reapplying the ML model to measure the impact on its output. If these features are actually the most relevant, their removal should lead to a significant change in the model's output, which, in the context of DR algorithms, is the projected layout.

We conducted an ablation study to evaluate the impact of removing the top-$k$ features identified by FADEx, Corbugy, LXDR, and ClusterShapley on the resulting DR layouts. The experiment uses the \href{https://archive.ics.uci.edu/dataset/33/dermatology}{Optical
 Recognition of Handwritten Digits} dataset (UCI ML Repository), which contains $5620$ instances and $64$ features. Its well-defined class structure facilitates interpretation and is particularly suitable for ClusterShapley, which relies on cluster information. We randomly sampled $10\%$ of the instances to reduce computational cost.

\begin{figure}[t]
    \centering
    \includegraphics[width=\columnwidth]{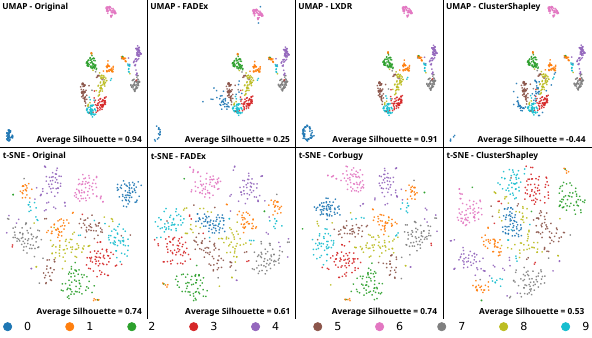}
    \caption{Ablation test. The left-most column presents the original projection made by UMAP and t-SNE. The other columns are the same embedding, but with the top-$15$ most important features ablated for class $0$. The bottom-right values are the average Silhouette scores (closer to $1$ is better) over 10 random seeds.}
    \label{fig:ablation}
\end{figure}

\cref{fig:ablation} shows the regular UMAP and t-SNE mappings (left-most) and their results after ablating the top-$15$ features from instances in class $0$. In the UMAP layouts (top row), ablating features identified by FADEx and ClusterShapley significantly disrupts the class $0$ cluster, pushing many points toward other classes. LXDR also affects the layout, but to a much lesser extent. A similar pattern appears in the t-SNE (bottom row): ablation with FADEx and ClusterShapley moves class $0$ toward the center of the layout, mixing it with class $8$, while with Corbugy’s method, class $0$ is shifted closer to classes $9$ and $5$ but largely preserving its structure and segregation.

To quantitatively assess the impact of the ablations derived from the four different DR explanation methods, we compared the Silhouette score (a metric that gauges cluster quality)~\cite{rousseeuw1987silhouettes} of cluster $0$ after applying the ablation. The values in the bottom-right of each plot in~\cref{fig:ablation} show the Silhouette scores. Notice that ablation from FADEx and ClusterShapley resulted in a substantial drop in the Silhouette (the closer to 1 the better) for both UMAP and t-SNE, corroborating the qualitative result discussed above.

Overall, the provided ablation study indicates that the top-ranked features identified by FADEx play a central role in shaping the embeddings. Notably, its behavior is comparable to ClusterShapley, despite the latter explicitly leveraging cluster information, highlighting FADEx’s effectiveness without requiring predefined clusters. \updated{A more comprehensive ablation evaluation is provided in the supplementary material.}

\subsubsection{The Feature Influence Vectors}

In this section, we demonstrate the effectiveness of FADEx’s Feature Influence Vectors in supporting the interpretation of how features shape the arrangement of instances in the visual space.

\begin{figure}
    \centering
    \includegraphics[width=\linewidth]{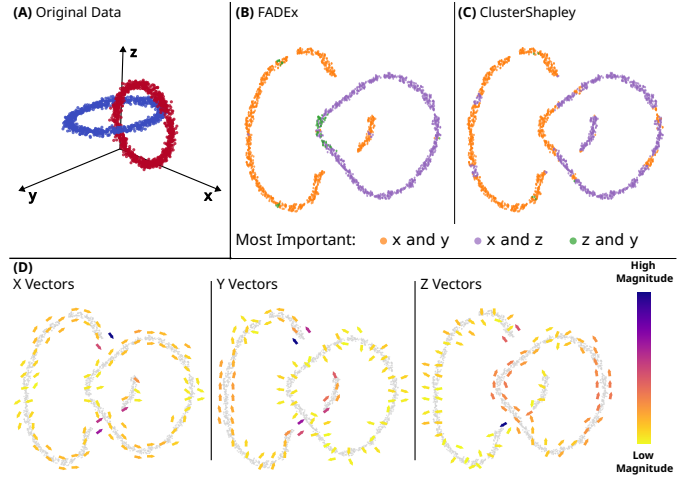}
    \caption{FADEx and ClusterShapley performance on snake-like cluster formations. (A) represents the high dimensional data. In (B) and (C), the low dimensional data was colored according to the most important feature for FADEx and ClusterShapley, respectively. In (D), FADEx's vector fields for each feature are plotted.}
    \label{fig:rings}
\end{figure}

Before presenting a qualitative analysis, notice from ~\cref{fig:rings}B) and C), which display the t-SNE (\texttt{perplexity=30}) embedding of the nested rings in ~\cref{fig:rings}A), colored according to the top-$2$ features identified by FADEx and ClusterShapley, respectively, that FADEx correctly identifies ${z,y}$ as the least relevant dimensions, as they rarely appear among the top-$2$ features for the blue and red ring structures. Although ClusterShapley also recognizes ${z,y}$ as less relevant, it incorrectly swaps the importance of ${x,y}$ and ${x,z}$ in the central region, as well as on the right of the embedding.

\cref{fig:rings}D) presents the Feature Influence Vectors computed by FADEx. From left to right, the vectors indicate the direction in which the embedding would shift under small perturbations along the $x$, $y$, and $z$ axes. The vectors become more tangential precisely in regions where the original 3D points align with the corresponding axis.
For instance, in the “X Vectors” (bottom left), the vectors follow the contour of the 2D point cloud along the top and bottom portions of the projected rings. These regions correspond to points located at the extremes of the $y$ (blue ring) and $z$ (red ring) axes in the original space. In contrast, vectors that appear nearly orthogonal to the point distribution indicate that perturbations in the corresponding feature would move points away from the main structure, i.e., perpendicular to the dominant grouping direction. A similar pattern is observed for the “Y Vectors” and “Z Vectors.” Overall, the Feature Influence Vectors provide an intuitive account of how individual features shape the layout arrangement. \updated{Additional experiments illustrating the usefulness of Feature Influence Vectors as an explanation resource is provided in Sec. 9  of supplementary material.}

\vspace{-0.15cm}

\section{Case Studies}
\label{sec:cs}

In this section, we present three case studies that highlight the usefulness and versatility of FADEx across different DR methods, and demonstrate how local feature attributions enable new forms of visual analytic reasoning about DR behavior.

\vspace{-0.15cm}

\subsection{Case Study 1}
\label{sec:cs1}

The first case study aims to demonstrate FADEx's ability to provide granular explanations for dimensionality reduction, highlighting \emph{which} features are the most important for each instance, \emph{where} and \emph{how} these features impact the projected layout. To do so, we applied our method to the \href{https://scikit-learn.org/stable/modules/generated/sklearn.datasets.load_breast_cancer.html}{Breast Cancer} dataset projected via UMAP, with default parameters. The dataset consists of $569$ instances and $30$ features describing characteristics of cell nuclei. 

\begin{figure}
    \centering
    \includegraphics[width=\columnwidth]{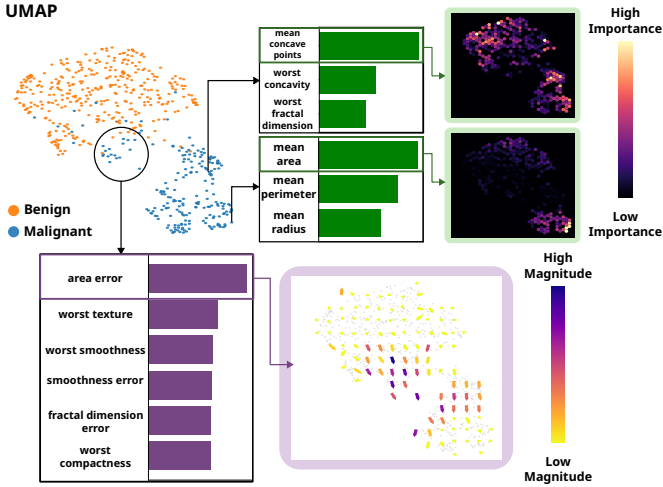}
    \caption{FADEx’s explanation of the UMAP embedding shows that \texttt{mean concave points} is important across multiple regions, while \texttt{mean area} has a more localized effect. The Feature Influence Vectors (bottom right) further reveal how features influence the layout. For example, perturbations in \texttt{area error} would push points toward the \texttt{Malignant} cluster, demonstrating that FADEx captures \emph{which} features matter, \emph{where}, and \emph{how} they affect the embedding.}
    \label{fig:cs1}
\end{figure}

The resulting projection, shown in ~\cref{fig:cs1} (top left), reveals two main clusters corresponding to \texttt{Benign} and \texttt{Malignant} instances. Although the \texttt{Malignant} instances are mapped separately from the \texttt{Benign} ones, they are not uniformly dense or cohesive. FADEx identifies \texttt{mean concave points} and \texttt{mean area} as the most important features for two instances located at the top and bottom of the \texttt{Malignant} cluster (top middle), respectively.
The heatmaps in the top right of ~\cref{fig:cs1} show that \texttt{mean concave points} plays an important role across multiple regions of the layout, whereas \texttt{mean area} is mainly relevant in the lower portion of the \texttt{Malignant} cluster. This highlights FADEx’s ability to capture localized semantic variations.

FADEx also proves effective in explaining transition zones. We selected a set of instances located at the boundary between the \texttt{Benign} and \texttt{Malignant} classes. For these transition points, FADEx identified \texttt{area error} as the most influential feature (bottom left). To understand the geometric impact of this feature, we analyzed its Feature Influence Vectors, which, at the boundary region, point directly toward the \texttt{Malignant} cluster. This directional information clearly indicates that an increase in \texttt{area error} would shift instances toward the malignant region. Such insights demonstrate that FADEx not only identifies \emph{which} features are important but also reveals \emph{where} and \emph{how} they impact the geometry of the reduced space.

\subsection{Case Study 2}
\label{sec:cs2}

\begin{figure*}[t]
    \centering
    \includegraphics[width=\linewidth]{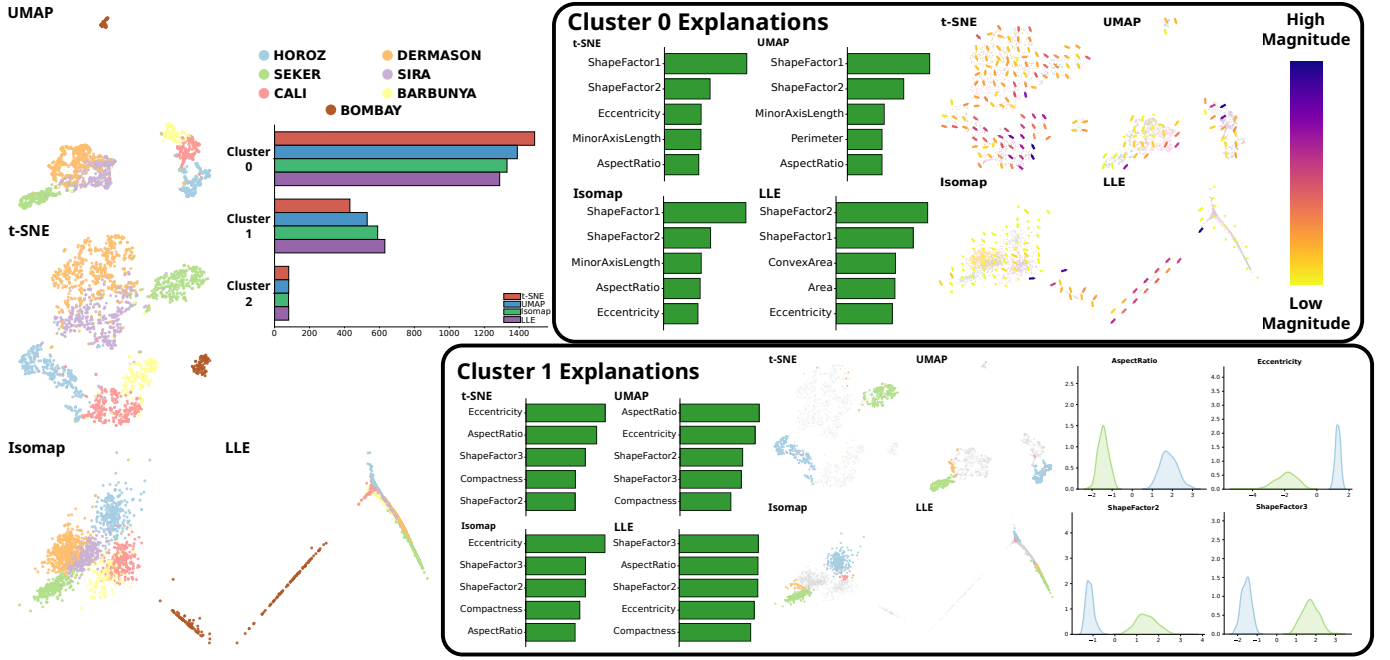}
    \caption{Comparing explanations across DR algorithms. Explanation vectors from all DR methods are clustered in the explanation space, and the middle-top bar plot shows their distribution by method. indicate strong agreement on feature importance across methods. In cluster $0$, the Feature Influence Vectors reveal method-specific use of \texttt{ShapeFactor1}. In cluster $1$, the four important features have a clearly separated distribution, corroborating FADEx's choice in pointing out those features as the most important.}
    \label{fig:cs2}
    \vspace*{-0.4cm}
\end{figure*}

In this case study, we exploit FADEx's model-agnostic approach to compare explanations derived from four DR algorithms, evaluating the extent to which the DR methods agree or disagree on the identification of the most relevant features and how a given feature affects the mapping of each DR method. In this study, we use the \href{https://archive.ics.uci.edu/dataset/602/dry+bean+dataset}{Dry Bean} dataset from the UCI repository, which consists of $13,611$ instances with $16$ features each. To facilitate visualization and reduce clutter, we selected a random sample of $2,000$ instances.

For each DR method, we apply FADEx to generate embeddings in the explanation space (\cref{eq:expl_space}). The feature attributions for each DR technique are normalized to the range $[0,1]$ for each \updated{feature}. The four sets of explanations are then clustered \updated{in the explanation space $E$} using $k$-means with $k=3$, after an optimum silhouette score search.

The scatter plots on the left in \cref{fig:cs2} show the layouts produced by t-SNE, UMAP, Isomap, and LLE. The bar plot in the top-middle reports the number of explanation vectors within each cluster. While the number of elements varies across DR methods for clusters $0$ and $1$, all four techniques exhibit the same pattern: most explanations concentrate in cluster $0$, followed by cluster $1$. This consistency suggests a shared understanding of feature importance across the mappings.
This agreement is further supported by the average Spearman correlation between explanation vectors across methods, with values ranging from $0.649$ for t-SNE and LLE to $0.842$ between t-SNE and UMAP.

The top-right and bottom panels in \cref{fig:cs2} show the averaged top-5 features for each DR method in clusters $0$ (top) and $1$ (bottom).
For cluster $0$ (top panel), there is clear agreement across methods: \texttt{ShapeFactor1} and \texttt{ShapeFactor2} consistently rank as the two most important features, with \texttt{MinorAxisLength} and \texttt{AspectRatio} figuring among the top-5 for t-SNE, UMAP, and Isomap. The Feature Influence Vectors for \texttt{ShapeFactor1} reveal that its effect varies by method. For t-SNE, larger displacements occur in the “Barbunya” and “Cali” classes, with similar directions, while a shift in direction appears when moving to the “Horoz” class. 
A direction shift is also observed when moving from “Sira” to “Dermason,” indicating that positive perturbations can tighten class groupings in those regions of the layout. UMAP shows a similar but less pronounced pattern, consistent with its more compact layout.
In contrast, Isomap exhibits vectors largely aligned in the same direction across classes, reflecting its more global behavior. For LLE, vectors have small magnitudes and highly variable directions, making interpretation difficult due to poor class separation, except for “Bombay,” where strong, aligned vectors suggest that perturbations in \texttt{ShapeFactor1} would further separate that region of the layout under positive perturbations.

Analyzing cluster $1$ in the bottom panel, we again observe strong agreement across methods, with features \texttt{AspectRatio}, \texttt{Eccentricity}, \texttt{ShapeFactor2}, and \texttt{ShapeFactor3} consistently appearing among the top-5 features for all DR methods. The scatterplots in the middle of the panel show that cluster $1$ is mainly composed of instances from the “Seker” and “Horoz” classes, which are well separated in all DR projections. The feature distribution plots on the right corroborate this separation: the four important features exhibit distinct distributions across the two classes, confirming that FADEx correctly identifies them as the most important features.

This case study shows that different DR methods tend to agree on the most important features, while Feature Influence Vectors reveal method-specific effects on the layout. These findings motivate further investigation into the generality of this pattern.

\subsection{Explaining Dimensionality Reduction for Images} 
\label{sec:cs3}

\begin{figure}[!t]
    \centering
    \includegraphics[width=\linewidth]{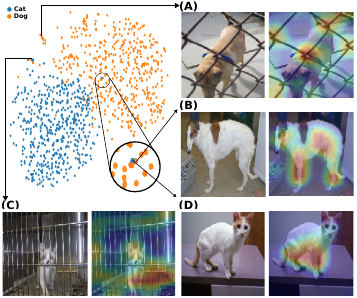}
    \caption{t-SNE embedding of the Cats and Dogs dataset with FADEx explanation maps. (A) and (C) show images from small upper-region clusters, where grid and wire fence features drive the projection. (D) shows a mispositioned cat in the dog cluster, with leg features dominating the embedding. (B) t-SNE also relies on leg features for the dog images, explaining their proximity with (D).}
    \label{fig:cats_dogs}
\end{figure}

This case study shows how FADEx's attribution scores can be employed to explain image data embeddings generated by DR methods. Specifically, we used a pre-trained VGG16~\cite{simonyan2014very} convolutional architecture as a feature extractor, originally pre-trained on the ImageNet dataset~\cite{imagenet}. We fine-tuned it using a sample of $1,000$ RGB images of cats and dogs ($500$ per class) from the Microsoft Cats and Dogs dataset~\cite{microsoft2017CatsDogs}. The images were resized to $224 \times 224$ pixels and normalized according to the ImageNet statistics to align with the pre-trained feature extractor.

The latent representation of each image is obtained from the output of the last convolutional layer (before the fully connected classification layer), which produces $512$ feature maps of size $7 \times 7$. We apply average pooling to each feature map, resulting in a $512$-dimensional feature vector representing the image’s latent embedding. 
We selected the last convolutional layer because it captures high-level semantic features and preserves visual representations of the input images.
The $512$-dimensional feature vectors are then projected onto a two-dimensional space using t-SNE, with a perplexity of $100$. 

In order to visualize the most influential features for each selected image, we identified the top $10$ features ranked by FADEx. The importance map is normalized to the interval $[0,1]$ and the \texttt{tensorflow} resize method is applied to generate a $224 \times 224$ map. The result is then overlaid on the original image as a heatmap to produce the feature importance visual representation.

\cref{fig:cats_dogs} shows the t-SNE layout with FADEx explanations highlighted for selected instances. The blue and orange clusters represent cats and dogs, respectively. Features corresponding to the faces of cats and dogs are identified as the most influential in determining the positioning of images in the projection space (see supplementary material, Sec. 8).
\cref{fig:cats_dogs}D) shows a mispositioned cat within the dog group. FADEx pointed out that the leg features contributed most to t-SNE positioning of the image, potentially introducing ambiguity that led to the image's incorrect placement. In \cref{fig:cats_dogs}B), we verify that t-SNE also used leg features to map a dog image very close to the image in \cref{fig:cats_dogs}D), which explains the proximity between the two images in the layout. 

FADEx explanations are also helpful in understanding why cat and dog images form the small clusters in the upper part of the layout (\cref{fig:cats_dogs}A) and C). The images in these clusters prominently contain grids and wire fences, which occupy a large portion of the image. During projection, t-SNE prioritizes features related to these elements, largely ignoring the animals themselves. As a result, images containing grids and fences are placed close to each other, regardless of whether they depict a cat or a dog.

This case study shows that FADEx identifies the key image regions driving the t-SNE embedding, linking visual structures to layout formation and providing insights into cluster patterns and DR behavior.
\section{Discussion and Limitations}

Explainable Dimensionality Reduction is an emerging field with many open challenges and opportunities for further research. The validation experiments in \cref{sec:validation} and the case studies in \cref{sec:cs} show that the FADEx framework goes beyond traditional local feature attribution, leveraging the SND distortion metric and the Feature Influence Vector fields to provide richer analysis. Specifically, FADEx not only uncovers \emph{which} features are most relevant for the mapping, but also \emph{where} distortions occur in the layout and \emph{how} the layout changes under feature perturbations. 
In particular, when dealing with data with a large number of \updated{features}, traditional approaches, such as analyzing the statistical distribution of each feature to understand cluster formation~\cite{thijssen2024interactive,yuan2022subplex}, can become impractical and ineffective.
FADEx streamlines the process by filtering for the top-$k$ relevant features, making the analytic process clearer and more efficient.
As the case studies demonstrate, this combination can unlock a wide range of new possibilities by providing deeper insights into how DR methods operate, facilitating the generation of new hypotheses, and improving trust in dimensionality reduction techniques across various domains.

As shown in \cref{sec:validation}, FADEx outperforms current state-of-the-art methods in both computational time and memory usage. In addition, it combines a set of properties not jointly found in existing approaches: model-agnosticism, assignment of a single score per feature, does not demand predefined clusters and out-of-sample mapping. In contrast, existing feature attribution methods such as Corbugy, LXDR, and ClusterShapley lack this combination of characteristics. By unifying feature attribution, Feature Influence Vector fields, and the SND distortion metric within a single framework, FADEx turns out to be a compelling and flexible tool for analyzing and interpreting DR layouts.

Applying FADEx to explain dimensionality reduction on image datasets is particularly compelling, as DR is widely used for exploring this type of data~\cite{Moharram2023}. The key features identified in the explanations shown in~\cref{fig:cats_dogs} appear to align closely with those used by the ML model for decision-making. While further investigation is required to confirm this relationship, our experiments open an interesting avenue for interpreting ML models through dimensionality reduction. 
Another interesting observation that emerged from our experiments, which deserves further analysis, is the substantial agreement in feature importance between different DR methods, particularly between UMAP and t-SNE. Does this agreement hold across a wider range of datasets?
Addressing these questions could provide valuable insights into different application scenarios. 

Regarding limitations, although we alleviate the challenges of high dimensionality through local PCA (\cref{sec:computational_aspects}), FADEx still depends on the quality of this dimensionality reduction step. When PCA fails to achieve a meaningful reduction without distorting the local structure of the data, the resulting approximations become less faithful, which can in turn compromise the reliability of the explanations.

\vspace{-0.2cm}

\section{Conclusions}

We presented FADEx, a model-agnostic local feature attribution method for explaining dimensionality reduction mappings. FADEx combines feature attribution, distortion assessment, and directional influence analysis within a unified framework based on local linear approximation via Taylor expansion and SVD, addressing key limitations of current DR explanation methods: it does not require out-of-sample data mapping, assigns a single interpretable attribution score per feature, works with any DR algorithm 
, and outperforms existing methods in computational efficiency. Our comprehensive evaluation across multiple datasets and DR methods demonstrates FADEx's effectiveness through quantitative metrics, comparisons with state-of-the-art methods, and diverse case studies. FADEx opens promising directions for future work, particularly integrating explanations into interactive visualization tools.

\acknowledgments{
	This work was supported by São Paulo Funding Agency (FAPESP), grants 2022/09091-8, 2025/10383-1 and 2025/18911-7; Brazilian National Council for Scientific and Technological Development (CNPq), grant 307184/2021-8. Ortigossa was supported by the Paulo Pinheiro de Andrade Fellowship. Silva was partially supported by the National Science Foundation grant OAC-2411221 and DARPA expMath program. Any opinions, findings, and conclusions or recommendations expressed in this material are those of the authors and do not necessarily reflect the views of the funding agencies and institutions they are affiliated with.}

\bibliographystyle{abbrv-doi-hyperref}

\bibliography{refs}

\appendix 
\crefalias{section}{appendix} 







\newpage

\begin{center}


\textbf{\Large Supplementary Material}
\end{center}

\section{FADEx Algorithm}

\begin{algorithm}[H]
    \caption{FADEx main pipeline (\texttt{fit} function)}
    \label{alg:fit}
\begin{algorithmic}[1]
    \REQUIRE $\mathbf{X} \in \mathbb{R}^{N \times n}$ (high-dimensional data), $\mathbf{Y} \in \mathbb{R}^{N \times k}$ (low-dimensional data), $i$ (index), \texttt{n\_neighbors} (neighborhood size)
    \ENSURE $\boldsymbol{\phi} \in \mathbb{R}^n$, $\text{SND} \in \mathbb{R}$, $J \in \mathbb{R}^{k \times n}$
    \STATE $\mathbf{x} \leftarrow \mathbf{X}[i]$; \quad $\mathbf{y} \leftarrow \mathbf{Y}[i]$; \quad $\mathbf{x}_0 \leftarrow \mathbf{x}$
    \STATE \COMMENT{Neighborhood selection}
    \IF{\texttt{n\_neighbors} is defined}
        \STATE $\mathcal{N}_x, \mathcal{N}_y \leftarrow \texttt{KNN}(\mathbf{X}, \mathbf{x}, \texttt{n\_neighbors})$
    \ELSE
        \STATE $\mathcal{N}_x \leftarrow \mathbf{X}$; \quad $\mathcal{N}_y \leftarrow \mathbf{Y}$
    \ENDIF
    \STATE \COMMENT{Dimensionality reduction preprocessing}
    \IF{\texttt{remove\_const\_feat}}
        \STATE $\mathbf{x}, \mathcal{N}_x, \texttt{mask} \leftarrow \texttt{drop\_const\_features}(\mathbf{x}, \mathcal{N}_x)$
    \ENDIF
    \IF{\texttt{use\_pca}}
        \STATE $\text{pca} \leftarrow \text{PCA}(0.95)$
        \STATE $\mathcal{N}_x \leftarrow \texttt{pca.fit\_transform}(\mathcal{N}_x)$; \quad $\mathbf{x} \leftarrow \texttt{pca.transform}(\mathbf{x})$
    \ENDIF
    \STATE \COMMENT{Jacobian estimation}
    \STATE $J \leftarrow \texttt{compute\_jacobian}(\mathbf{x}, \mathbf{y}, \mathcal{N}_x, \mathcal{N}_y, \lambda=0.1)$
    \STATE \COMMENT{Restore full feature space}
    \IF{\texttt{use\_pca}}
        \STATE $J \leftarrow J \cdot \mathbf{W}_{\mathbf{x}}$
    \ENDIF
    \IF{\texttt{remove\_const\_feat}}
        \STATE $J \leftarrow \texttt{reinflate\_jac}(J, \texttt{mask})$
    \ENDIF
    \STATE \COMMENT{Importance and distortion}
    \STATE $\boldsymbol{\phi} \leftarrow \texttt{compute\_importance}(J, \mathbf{x}_0)$; \quad $\text{SND} \leftarrow \| J \|_{spec}$
    \RETURN $\boldsymbol{\phi},\; \text{SND},\; J$
\end{algorithmic}
\end{algorithm}


    


    
    

    



    

    

    
\begin{algorithm}[H]
    \caption{\texttt{compute\_jacobian}}
    \label{alg:compute_jacobian}
\begin{algorithmic}[1]
    \REQUIRE $\mathbf{x} \in \mathbb{R}^n$, $\mathbf{y} \in \mathbb{R}^k$, $\mathcal{N}_x \in \mathbb{R}^{\texttt{n\_neighbors} \times n}$, $\mathcal{N}_y \in \mathbb{R}^{\texttt{n\_neighbors} \times k}$, $\lambda$
    \ENSURE $J \in \mathbb{R}^{d \times D}$
    \STATE \COMMENT{Local displacements}
    \STATE $\Delta X \leftarrow \mathcal{N}_x - \mathbf{x}^\top$; \quad $\Delta Y \leftarrow \mathcal{N}_y - \mathbf{y}^\top$
    \STATE \COMMENT{Gaussian distance weights}
    \STATE $r \leftarrow \| \Delta X \|_2, \quad \ell = 1, \ldots, \texttt{n\_neighbors}$
    \STATE $\sigma \leftarrow \text{median}(\{r : r > 0\})$
    \STATE $w \leftarrow \exp\!\left( -\dfrac{r^2}{2\,\sigma^2} \right), \quad \ell = 1, \ldots, \texttt{n\_neighbors}$
    \STATE \COMMENT{Solve the weighted ridge system and transpose to Jacobian shape}
    \STATE $A \leftarrow \Delta X^\top \, \text{diag}(\mathbf{w}) \, \Delta X + \lambda\, I_D$
    \STATE $C \leftarrow \Delta X^\top \, \text{diag}(\mathbf{w}) \, \Delta Y$
    \STATE $B \leftarrow \texttt{numpy\_solver}(A, C)$
    \STATE $J \leftarrow B^\top$
    \RETURN $J$
\end{algorithmic}
\end{algorithm}

    
    
    


    
    

\begin{algorithm}[H]
    \caption{\texttt{drop\_const\_features}}
    \label{alg:drop_const_features}
\begin{algorithmic}[1]
    \REQUIRE $\mathbf{x} \in \mathbb{R}^n$, $\mathcal{N}_x \in \mathbb{R}^{\texttt{n\_neighbors} \times n}$, $\tau = 10^{-6}$
    \ENSURE $\mathbf{x}', \mathcal{N}_x'$, \texttt{mask}
    \STATE $\boldsymbol{\sigma}^2 \leftarrow \operatorname{Var}(\mathcal{N}_x, \text{axis}=0)$
    \STATE $\texttt{mask} \leftarrow \boldsymbol{\sigma}^2 > \tau$
    \STATE $\mathbf{x}' \leftarrow \mathbf{x}[\texttt{mask}]$
    \STATE $\mathcal{N}_x' \leftarrow \mathcal{N}_x[\texttt{mask}]$
    \RETURN $\mathbf{x}', \mathcal{N}_x', \texttt{mask}$
\end{algorithmic}
\end{algorithm}

    

    
    
    
    
    
    

\begin{algorithm}[H]
    \caption{\texttt{compute\_importance}}
    \label{alg:compute_importance}
\begin{algorithmic}[1]
    \REQUIRE $J \in \mathbb{R}^{k \times n}$, $\mathbf{x} \in \mathbb{R}^n$
    \ENSURE $\boldsymbol{\phi} \in \mathbb{R}^n$
    \STATE $\mathbf{U}, \Sigma, \mathbf{V}^T \leftarrow \texttt{svd}(J)$
    \STATE $\lambda_1, \lambda_2 \leftarrow \Sigma_{11}, \Sigma_{22}$
    \FOR{$j = 1, \ldots, n$}
        \STATE $\phi_j \leftarrow |V_{1j} \cdot x_j| + \frac{\lambda_2}{\lambda_1} \cdot |V_{2j} \cdot x_j|$
    \ENDFOR
    \RETURN $\boldsymbol{\phi}$
\end{algorithmic}
\end{algorithm}

    

    
    
    
    

\begin{algorithm}[H]
    \caption{\texttt{reinflate\_jac}}
    \label{alg:reinflate_jac}
\begin{algorithmic}[1]
    \REQUIRE $J_{\text{filt}} \in \mathbb{R}^{k \times n'}$ (filtered Jacobian), \texttt{mask} $\in \{0,1\}^n$ (boolean mask over original features)
    \ENSURE $J_{\text{full}} \in \mathbb{R}^{k \times n}$
    \STATE $J_{\text{full}} \leftarrow \mathbf{0}^{k \times n}$
    \STATE $J_{\text{full}}[:,\; \texttt{mask}] \leftarrow J_{\text{filt}}$
    \RETURN $J_{\text{full}}$
\end{algorithmic}
\end{algorithm}

\section{Computing Jacobian Using the Chain Rule}

As discussed in Section 3.2 of the main manuscript, we can apply local PCA to reduce the dimensionality of the data. The Jacobian matrix is then estimated in the PCA reduced space and then mapped to the original space using the chain rule for derivatives. In the following, we detail this mathematical construction.

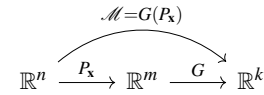
\begin{figure}[!h]
    \centering
    $$
\begin{tikzcd}
\mathbb{R}^n 
  \arrow[r,"P_{\mathbf{x}}"] 
  \arrow[rr,bend left=40,"\mathcal{M}=G( P_{\mathbf{x}})"]
& \mathbb{R}^m 
  \arrow[r,"G"] 
& \mathbb{R}^k
\end{tikzcd}
 $$
    \caption{Decomposing the DR mapping as a combination of PCA projection $P$ and an auxiliary mapping $G$.}
    \label{fig:mappings}
\end{figure}

Consider the following mappings as illustrated in Fig.~\ref{fig:mappings}: 
\begin{enumerate}
    \item $\mathcal{M}: \mathbb{R}^n \to \mathbb{R}^k$, $k<<n$, the dimensionality reduction (DR) method we intend to explain (typically $k=2$),
    \item $P_{\mathbf{x}}:\mathbb{R}^n \to \mathbb{R}^m$, $m<n$, the local PCA projection (in the neighborhood of ${\mathbf{x}}$),
    \item $G:\mathbb{R}^m \to \mathbb{R}^k$ a mapping that relates points in the PCA space with their image in $\mathbb{R}^k$ generated by $\mathcal{M}$.
\end{enumerate}

By the chain rule, the Jacobian of $\mathcal{M}$ at a point $\mathbf{x} \in \mathbb{R}^n$ is:
\begin{equation}
    J_{\mathcal{M}(\mathbf{x})} = J_G(P_{\mathbf{x}}) \cdot J_{P_{\mathbf{x}}}
\end{equation}
where:
\begin{itemize}
    \item $J_{\mathcal{M}(\mathbf{x})}  \in \mathbb{R}^{k \times n}$ is the sought Jacobian,
    \item $J_G(\mathbf{z}) \in \mathbb{R}^{k \times m}$ is the Jacobian in the reduced PCA space, with $\mathbf{z} = P_{\mathbf{x}}$,
    \item $J_{P_{\mathbf{x}}} \in \mathbb{R}^{m \times n}$ is the Jacobian of the PCA projection.
\end{itemize}

Since $P_{\mathbf{x}}(\mathbf{u})=\mathbf{W}_{\mathbf{x}}(\mathbf{u})$, where $\mathbf{W}_{\mathbf{x}}$ is the local  PCA projection matrix, we have that $J_{P_{\mathbf{x}}}=\mathbf{W}_{\mathbf{x}}$. Therefore:

$$
J_{\mathcal{M}(\mathbf{x})}= J_g(\mathbf{z})\cdot\mathbf{W}_{\mathbf{x}}
$$

\section{Comparing FADEx's distortion metric (SND) with Hypertrix}

\begin{figure}[H]
    \centering
    \includegraphics[width=1\linewidth]{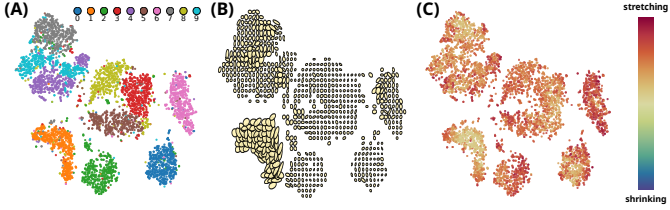}
    \caption{(A) Projected space produced by t-SNE on the MNIST dataset. (B) and (C) Hypertrix ellipse and FADEx SND distortion plots, respectively. Notice the clear disagreement between SND and Hypertrix for the digit 1 cluster. }
    \label{fig:snd_hypertrix_sm}
\end{figure}

Hypertrix is a method for representing distortions created by dimensionality reduction (DR) algorithms. The method decomposes the embedding space into a regular grid, using points inside and in the neighborhood of each grid cell to build a linear transformation $\mathbf{A}$, that locally approximates the DR algorithm. The eigen vectors and values of $ Cov = \mathbf{A}^\top \mathbf{A}$ are used to draw an ellipse centered in each grid cell. Ellipses are colored based on the ratio $\delta = D / d$, where $D$ and $d$ are average distances in high-dimensional and projected spaces, respectively.

Unlike Hypertrix, FADEx approximates one linear transformation per point, instead of one per grid cell, providing a distortion assessment with finer granularity. 

To provide a fair comparison, we performed the same experiment proposed in Hypertrix's paper with MNIST. We mapped $5000$ samples of the dataset using t-SNE (\texttt{perplexity=30}, \texttt{early\_exaggeration=24}, \texttt{learning\_rate=600}, \texttt{'pca'}) and computed both FADEx and Hypertrix distortions. The analysis of FADEx performance in this dataset is addressed in Section 4.2 of the main manuscript. Fig.~\ref{fig:snd_hypertrix_sm} presents the projection space, produced by t-SNE, the Hypertrix and FADEx distortion plots. Comparing Fig.~\ref{fig:snd_hypertrix_sm}(B) and Fig.~\ref{fig:snd_hypertrix_sm}(C), there is a clear disagreement between Hypertrix and FADEx in relation to the digit 1 cluster: while FADEx's SND metric shows low distortion, Hypertrix ellipses are big and stretched, indicating a region of high distortion. After a meticulous analysis over Hypertrix algorithm, we identified mathematical issues and an inconsistency between the code provided in the supplementary material and the algorithm described in the paper.

Regarding the inconsistency, Hypertrix paper defines the ellipses axis as $a = 1/\sqrt{\lambda_1}$ and $b = 1/\sqrt{\lambda_2}$, while its code computes: $a = \sqrt{\lambda_1}$ and $b = \sqrt{\lambda_2}$, where $\lambda_1$ and $\lambda_2$ are singular values of $Cov = A^\top A$. We assume this is just a typo mistake from the paper, since the code implementation makes much more sense, as the singular values quantify the distortion, without the need of inversion.

In relation to the mathematical issues, we identified that the Hypertrix's algorithm: (1) applies a global PCA, when the dimensionality surpasses 50; (2) does not compute the original linear transformation via Chain Rule; and (3) it gathers neighbors in the projected space.

\begin{itemize}
    \item \textbf{1. Global PCA:} This algorithm discards directions in feature space that capture little variance across the entire dataset. However, a direction with low global variance can exhibit high local variance within a specific local neighborhood, and discarding it may spoil important local information in the data.

    \item \textbf{2. No Chain Rule:} By not applying the Chain Rule, Hypertrix is computing a mapping $G$ (see Fig.~\ref{fig:mappings}) that does not approximate the DR algorithm, but rather generates a mapping from the PCA space to the visual space without accounting for the original features.

    \item \textbf{3. Neighborhood in the Projected Space:} By searching for neighbors in the t-SNE space, Hypertrix inherits the distortion produced by the DR algorithm. As we showed in Section 4.2 of the main manuscript, the true neighbors in high dimension can be mapped far apart in the projected space. It means that the neighbors in the projected space may not be truly neighbors, and using them to approximate a linear transformation can make the matrix ill-conditioned. 
\end{itemize}

When looking at the disagreement between Hypertrix and FADEx, with these issues in mind, it makes sense to assume that the global PCA discarded directions with local high variance for digit 1, compacting it in the PCA space, and the linear transformation learned by Hypertrix maps this compacted cluster to the stretched formation in the projected space. In this sense, the ellipses plotted by Hypertrix tell little about the t-SNE transformation, while FADEx's SND metric actually assesses the t-SNE distortion.


\section{Comparing FADEx's Influence Vector Fields with FeatureClock Biplots}

FeatureClock is a visualization method that explains non-linear dimensionality reduction by fitting linear regression models between the original features and the projection of the 2D embedding onto lines at varying angles. It provides three types of visualizations:

\begin{itemize}
    \item \textbf{1. Global Clock:} Fits a single linear model over all data points to identify the direction of each features's strongest contribution in the embedding.
    \item \textbf{2. Local Clocks:} Apply the same procedure within individual clusters to capture finer-grained, group-level feature effects
    \item \textbf{3. Inter-group Clock:} Uses logistic regression to identify features that discriminate between user-defined groups displaying contributions along the axis connecting cluster centers.
\end{itemize}

\begin{figure}[H]
    \centering
    \includegraphics[width=1\linewidth]{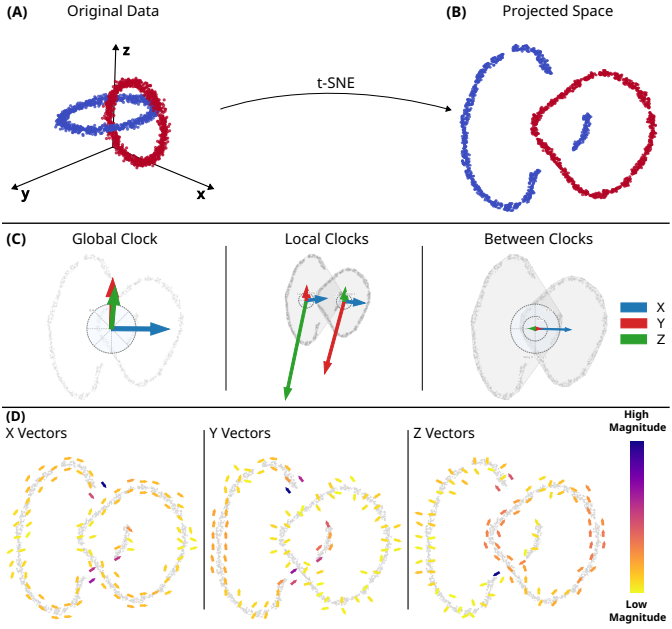}
    \caption{(A) Interlocked rings artificial dataset visualized in its original 3D space. (B) projected space produced by t-SNE. (C) FeatureClock biplots for this dataset. (D) FADEx Influence Vector Fields for the same dataset.}
    \label{fig:vectors_sm}
\end{figure}

To perform the comparison with FADEx, we used the same experiment from Section 4.3.2 of the main manuscript. Fig.~\ref{fig:vectors_sm}(A) and (B) present the same high- and low-dimensional space of such an experiment.

Fig.~\ref{fig:vectors_sm}(C) shows the three FeatureClock visualizations for the interlocked rings dataset embedded by t-SNE. None of the clocks could actually capture each feature's influence across the dataset. The Global Clock assigns large uniform contributions to Y and Z that do not reflect their selective relevance to each ring's local structure. For the Local Clocks, the oversized arrows represent the dominant linear axis variation within each ring, rather than the true directional influence of those features along the curved structure. The Between Clock produces small arrows, since the interlocked geometry prevents effective linear separation between the two classes.

In contrast, FADEx's Influence Vector Fields in Fig~\ref{fig:vectors_sm}(D) correctly reveal that X exhibits high magnitude vectors across both rings, while Y and Z vectors are concentrated on the blue and red rings, respectively. This is consistent with the ground-truth construction where the blue ring lies in the XY plane and the red ring in the XZ plane.

\updated{

\section{Visualization-Encoding Choices}

FADEx is designed so that attribution, distortion, and directional influence all derive from a single per-instance local linear operator. This unified approach has a direct consequence for visualization: the three quantities can be displayed through encodings that are already familiar to practitioners of feature-attribution methods (bar charts for per-feature importance, scalar color maps for the SND distortion, and arrow glyphs for the Feature Influence Vectors) rather than requiring a specialized visual idiom. This design choice reduces the cognitive burden on users by leveraging visualization conventions they are likely already familiar with, making the layouts easier to interpret.

Specialized encodings developed for DR explanation, such as the indicatrix glyphs of Hypertrix, the axis lines of DimReader, and the clock metaphor of FeatureClock, can convey richer per-point geometric information in a single mark. However, they typically require the reader to learn a new visual vocabulary and are tailored to a specific aspect of the explanation (e.g., local distortion for Hypertrix, axis-aligned sensitivity for DimReader). FADEx instead exposes the underlying operator, from which each of these aspects can be read out separately using standard, low-overhead encodings. This keeps the individual visualizations simple and composable, at the cost of not packing all geometric information into a single dense glyph.

We note that the encodings used in this paper are not the only ones compatible with FADEx. Because the method yields different explanation resources per instance, such information can, in principle, drive more advanced visual encodings developments (including indicatrix-style glyphs analogous to those of Hypertrix). 

\section{Further Assessment of the Approximation Quality}

As discussed in section 3.3 of the main manuscript, we evaluated the approximation quality of the Jacobian estimator with a hold-out protocol using the mean cosine similarity metric. In this section of the supplementary material, we aim to better justify the use of such a metric, as well as present a new evaluation using an additional metric on different datasets and DR algorithms.

Non-linear DR methods such as t-SNE and UMAP produce embeddings whose overall scale is arbitrary, depending on factors such as initialization, learning rate, and the number of optimization iterations. Scale-invariant measures are therefore necessary for assessing and comparing quality scores across different DR methods and datasets. The mean cosine similarity is a suitable metric for such evaluation, since it is scale-invariant and, by assessing the angular alignment between the predicted and observed displacement vectors, it captures whether the estimated operator places test points (held-out) in the correct direction within the embedding.

While the cosine similarity captures directional agreement, it is by construction insensitive to the magnitude of the displacements: an operator could predict the correct directions while systematically over- or under-estimating how far points move in the embedding. To assess this complementary aspect, we additionally evaluate the Pearson correlation between the predicted and observed displacement magnitudes over the held-out neighbors. Like the cosine similarity, the Pearson correlation is scale-invariant and therefore remains comparable across different datasets. A high correlation indicates that the estimated Jacobian operator preserves the relative scale of local displacements (that larger displacements in the original mapping are mapped to proportionally larger displacements by the Jacobian). This is particularly relevant because the magnitude of the displacements is governed by the singular values of the Jacobian operator, which scale the projection along each singular direction, affecting both the feature attribution scores and the SND distortion measure.

Fig.~\ref{fig:approx_quality_sm} reports the mean cosine similarity and the Pearson correlation as a function of the neighborhood size for three datasets, each embedded with four DR methods. The most consistent trend across all panels concerns the influence of the DR algorithm's own parameter (\texttt{perplexity} for t-SNE and \texttt{n\_neighbors} for others): higher parameter values yield uniformly higher approximation quality for both metrics, whereas lower values produce lower quality approximations. This confirms that smoother embeddings, induced by larger DR parameter values, are more faithfully captured by the local linear approximation, consistent with the discussion in Sec. 3.3. 

The effect of the neighborhood size, by contrast, depends strongly on the nature of the dataset. For the Breast Cancer dataset, both metrics increase with the neighborhood size and then either stabilize or continue to grow only marginally. For the Musk Version 2 dataset, both metrics rise sharply for small neighborhoods and reach a maximum at a moderate neighborhood size, after which they either decline or plateau. For the Swiss Roll dataset, the approximation quality is highest at small neighborhoods and usually degrades as the neighborhood grows, since enlarging it quickly extends the fit beyond the locally linear regime.

These observations have a direct implication for the choice of the neighborhood size in FADEx. In all cases, there exists a neighborhood size at which both metrics are higher; beyond it, the approximation quality either declines, as in the Swiss Roll dataset, or improves so marginally that the gain does not justify the accompanying loss of locality, as in the Breast Cancer dataset. The location and sharpness of this optimum depend on the nature of the dataset and its hyperparameter value, as seen across the panels in Fig.~\ref{fig:approx_quality_sm}. For datasets that yield smoother embeddings and are densely sampled relative to their dimensionality, such as Breast Cancer, the neighborhood size has little effect on the approximation quality over a wide range; in this regime, k acts less as a fidelity parameter than as a choice of the spatial scope from which the explanation is extracted, leaving the user freedom to select the neighborhood size they wish to account for. For datasets embedded through more strongly non-linear mappings, such as the Swiss Roll, the approximation is faithful only within a well-defined optimal neighborhood, which is typically small, since enlarging it rapidly violates the local-linearity assumption. Empirically, across the datasets considered, we observed that the optimal neighborhood size is around $10\%$ of the dataset size. This should not be interpreted as a strict recommendation, but rather as an empirical indication of a good trade-off between locality and approximation quality.

\newpage

\begin{figure*}[p]
  \centering
  \includegraphics[width=\textwidth, height=0.88\textheight]{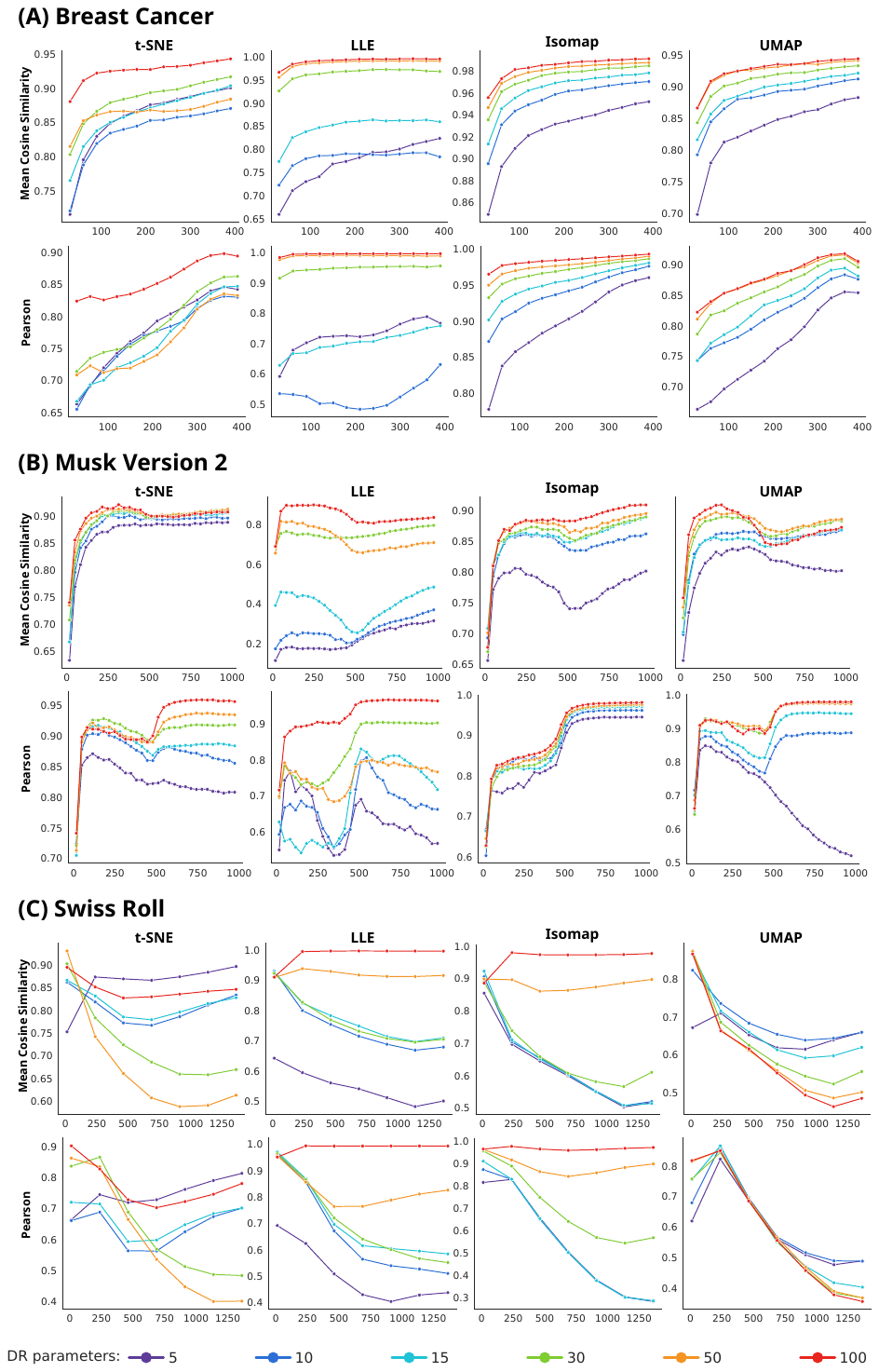}
  \caption{(A), (B) and (C) present the mean cosine similarity (top row) and Pearson correlation (bottom row) between predicted and observed displacements as a function of the neighborhood size, evaluated under the hold-out protocol of Section 3.3 of the main manuscript, for the Breast Cancer, Musk Version 2 and Swiss Roll dataset, respectively. Breast Cancer and Musk Version 2 are as described in the main manuscript and Swiss Roll was generated with the \href{https://scikit-learn.org/stable/modules/generated/sklearn.datasets.make_swiss_roll.html}{scikit-learn function}, using $2000$ samples and $0.1$ noise.}
  \label{fig:approx_quality_sm}
\end{figure*}

\newpage

\section{Extended Ablation Analysis}

We extended the ablation experiment of Section ~4.3.1 from the main manuscript into a systematic sweep over the number $m$ of ablated features, ranging $m$ from $1$ to $15$. For each value of $m$, we ablate the top-$m$ features ranked by each method and compute the Silhouette score of the affected cluster, allowing us to assess whether FADEx holds higher quality across ablation levels rather than at a single fixed value (see Figure~\ref{fig:ablation_sweep}). 
In addition, we replicated the experiment of Section~4.3.1 on the Dermatology dataset (described in the main manuscript), ablating features from the \textit{seborreic and chronic dermatitis} clusters. The resulting embeddings when ablating the \textit{seborreic dermatitis} are shown in Figure \ref{fig:dermatology_ablation}. 

Figure~\ref{fig:ablation_sweep} reports the silhouette of the ablated clusters as a function of $m$, averaged over five seeds, for two clusters of each dataset and for both UMAP and t-SNE. Across all four clusters, ablating the top features ranked by FADEx degrades the cluster structure substantially more than removing random features, confirming that the features FADEx ranks highest are genuinely those that sustain the cluster in the embedding, and that this behavior holds across ablation levels rather than at a single fixed $m$. Under UMAP, the effect is most pronounced and comparable to ClusterShapley, despite the latter relying on predefined cluster information, while LXDR, Corbugy, and the random baseline leave the cluster largely intact. The same trend is observed under t-SNE, although the curves are noisier because t-SNE is non-parametric and provides no fixed mapping to project ablated points; each ablation requires re-running the embedding, which introduces additional variability. We note that for large $m$ the curves can become non-monotonic, as removing a large fraction of the features reorganizes the layout globally; this is visible, for instance, in the UMAP \textit{chronic dermatitis} panel, where the FADEx silhouette begins to rise again at large $m$. The small-to-moderate $m$ regime is therefore the most informative.

\begin{figure}[H]
    \centering
    \includegraphics[width=\linewidth]{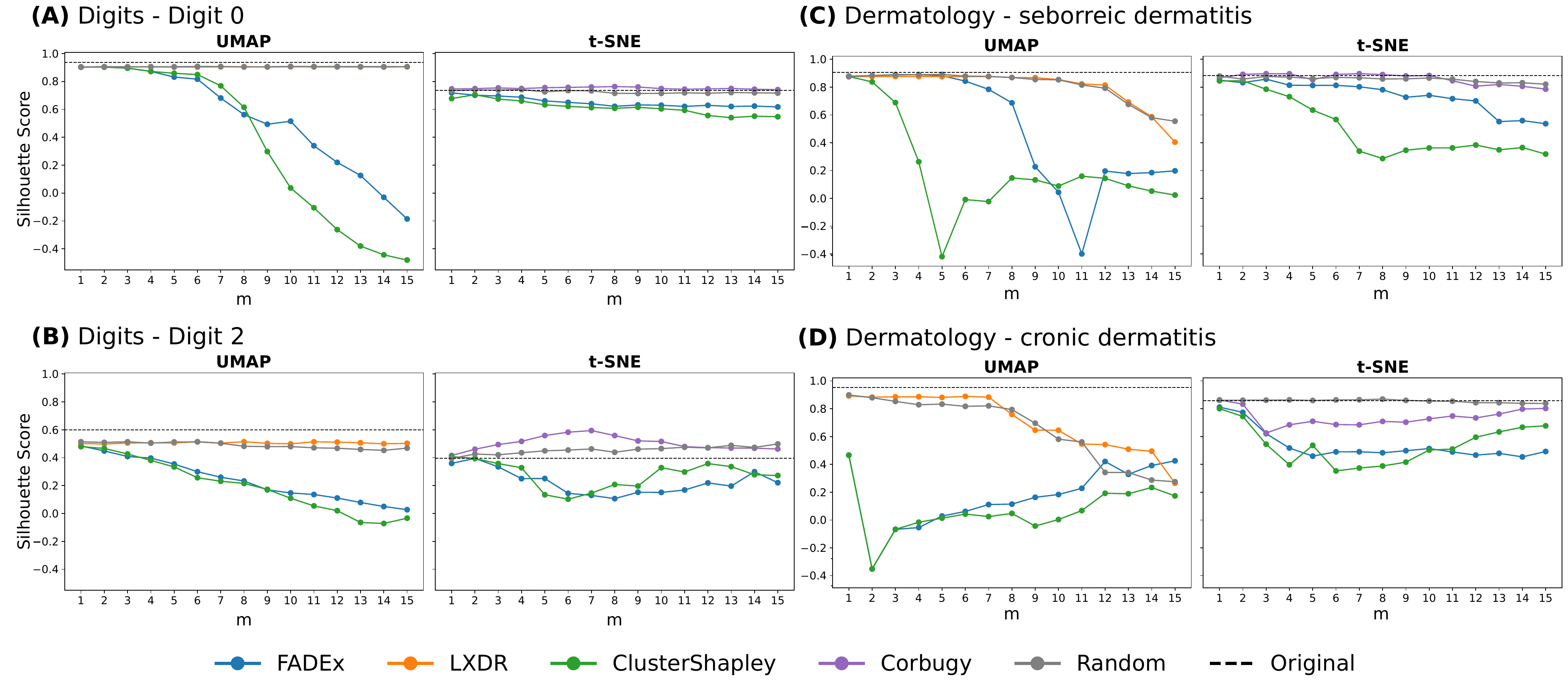}
    \caption{Ablation sweep: silhouette of the ablated cluster as a function of the number of ablated features $m$ (from $1$ to $15$), averaged over five seeds, for UMAP (left) and t-SNE (right) in each panel. Each curve corresponds to ranking the ablated features with a different method: FADEx, ClusterShapley, LXDR (UMAP only), Corbugy (t-SNE only), and a random-ablation baseline; the dashed line indicates the original silhouette before ablation. (A) Digits, class~0; (B) Digits, class~2; (C) Dermatology, \textit{seborreic dermatitis}; (D) Dermatology, \textit{chronic dermatitis}. Across all clusters, ablating FADEx-ranked features degrades the cluster more than removing random or LXDR-ranked features. Curves for large $m$ are less informative, as ablating many features reorganizes the layout globally. We selected these classes because their original silhouette scores were high, providing a clear baseline against which the effect of ablation can be measured.}
    \label{fig:ablation_sweep}
\end{figure}

\newpage

\begin{figure}[H]
    \centering
    \includegraphics[width=\linewidth]{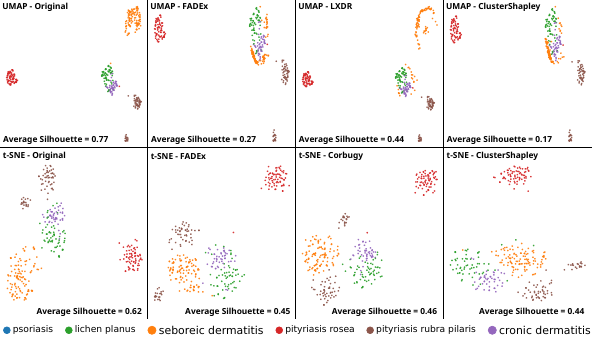}
    \caption{Ablation experiment of Section~4.3.1 applied to the Dermatology dataset, ablating for the \textit{seborreic dermatitis} cluster. Top row: UMAP layouts; bottom row: t-SNE layouts. As in the main manuscript, ablating FADEx-ranked features substantially reduces the silhouette, comparable to ClusterShapley.}
    \label{fig:dermatology_ablation}
\end{figure}

}

\newpage
\section{Explaining Dimensionality Reduction for Images}

\begin{figure}[H]
    \centering
    \includegraphics[width=1\linewidth]{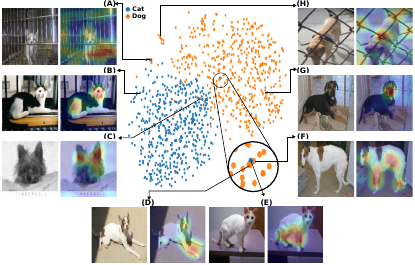}
    \caption{t-SNE mapping of 512-dimensional features extracted from 1,000 images (500 cats and 500 dogs) using a VGG16
    convolutional network, together with explanation maps generated by FADEx. \textbf{(A) and (H)}: Small clusters generated by t-SNE in the top-left part of the projected space. FADEx reveals that the DR algorithm used fences and wire features to embed the image, which explains why they are isolated. \textbf{(B) and (G):} Images correctly mapped, t-SNE used the dog and cat's features (mainly from the face) to perform the mapping. \textbf{(C):} a mispositioned dog image in the transition zone of both classes. The image has poor quality, and the DR algorithm used the dog's cat-like ears to embed the image, explaining why the point is surrounded by cats. \textbf{(D), (E) and (F):} Three images are very closely mapped, including a misplaced cat. FADEx explains that t-SNE used the animal's leg features to embed the image. }
    \label{fig:cats_dogs_sm}
\end{figure}

\updated{

\newpage
\section{Feature Influence Vectors on the Swiss Roll dataset}

\begin{figure}[H]
    \centering
    \includegraphics[width=1\linewidth]{figs/swiss_roll_sm.pdf}
    \caption{FADEx Feature Influence Vectors on the \href{https://scikit-learn.org/stable/modules/generated/sklearn.datasets.make_swiss_roll.html}{Swiss-roll dataset}, illustrating the method's behavior on a continuous manifold. (A) The original 3D Swiss-roll, colored by position along the rolled axis, with $2000$ points. (B) The t-SNE embedding, which tears the manifold into two bands. (C) Feature Influence Vectors for the $x$, $y$, and $z$ axes, overlaid on a grid sample in the embedding space. Each arrow indicates the direction in which a point would move under a small positive perturbation of the corresponding feature. The $y$ vectors exhibit the largest, most coherent magnitudes, reflecting that $y$ (the axis along which the manifold rolls) most strongly drives the embedding, whereas the $x$ and $z$ vectors are smaller and follow the local contour of each band. Unlike the two-rings example from the main manuscript, no single feature dominates a region of the layout. The Feature Influence Vectors nonetheless provide an interpretable, continuously varying account of how each feature shapes the embedding.}
    \label{fig:swiss_roll_vectors}
\end{figure}

\newpage
\section{Computation Cost Comparison Table}

\begin{table}[H]
    \centering
    \caption{Computational times and memory usage comparison. Metrics represent the mean of 10 independent runs for execution time and the peak Resident Set Size (RSS) increment for memory. Each run corresponds to computing the explanation for a single instance. FADEx and LXDR used the same neighborhood size.}
    \label{tab:cost_comparison}
    \small
    \setlength{\tabcolsep}{5pt}
    \renewcommand{\arraystretch}{1.2}
    
    \begin{tabular}{|l|l|@{\hspace{2pt}}c@{\hspace{2pt}}|@{\hspace{2pt}}c@{\hspace{2pt}}|@{\hspace{2pt}}c@{\hspace{2pt}}|@{\hspace{2pt}}c@{\hspace{2pt}}|}
        \hline
        \textbf{Dataset} & \textbf{Metric} & \textbf{FADEx} & \textbf{LXDR} & \textbf{ClusterShapley} & \textbf{Corbugy} \\
        \hline
        \hline
        
        \multirow{2}{*}{%
            \begin{tabular}{@{}l@{}}
                \href{https://archive.ics.uci.edu/dataset/17/breast+cancer+wisconsin+diagnostic}{\textbf{Breast Cancer}}
            \end{tabular}%
        } 
        & \textbf{Time (s)}    & $0.02$ & $0.19$ & $0.13$ & $0.02$ \\
        & \textbf{Memory (MB)} & $6.42$ & $9.96$ & $16.28$ & $9.26$ \\
        \hline
        
        \multirow{2}{*}{%
            \begin{tabular}{@{}l@{}}
                \href{https://archive.ics.uci.edu/dataset/75/musk+version+2}{\textbf{Musk Version 2}} \\
                ($2000$ samples)
            \end{tabular}%
        }   
        & \textbf{Time (s)}    & $0.07$ & $1.38$ & $0.53$ & $0.28$ \\
        & \textbf{Memory (MB)} & $12.43$ & $19.66$ & $58.20$ & $64.61$ \\
        \hline
        
        \multirow{2}{*}{%
            \begin{tabular}{@{}l@{}}
                \href{https://scikit-learn.org/stable/modules/generated/sklearn.datasets.fetch_openml.html}{\textbf{MNIST}} \\
                (5000 samples)
            \end{tabular}%
        }      
        & \textbf{Time (s)}    & $0.46$ & $1.46$ & $7.25$ & $2.10$ \\
        & \textbf{Memory (MB)} & $70.16$ & $137.46$ & $279.99$ & $475.45$ \\
        \hline
        
        \multirow{2}{*}{%
            \begin{tabular}{@{}l@{}}
                \href{https://archive.ics.uci.edu/dataset/54/isolet}{\textbf{Isolet}} \\
                (5000 samples)
            \end{tabular}%
        }         
        & \textbf{Time (s)}    & $0.41$ & $0.72$ & $7.88$ & $1.88$ \\
        & \textbf{Memory (MB)} & $41.23$ & $67.85$ & $306.84$ & $456.27$ \\
        \hline
        
    \end{tabular}
\end{table}

}








\end{document}